%% file: reforce_nips25.tex
\documentclass{article}

\input{math_commands.tex}

\usepackage[numbers]{natbib}
\usepackage{graphicx}
\usepackage{listings}
\usepackage{booktabs}
\usepackage{multirow}
\usepackage{enumitem}
\usepackage{subcaption}
\usepackage{caption}
\usepackage{framed}
\usepackage{fancybox}
\usepackage[most]{tcolorbox}
\usepackage{tabularx}
\colorlet{punct}{red!60!black}
\definecolor{background}{HTML}{EEEEEE}
\definecolor{delim}{RGB}{20,105,176}
\colorlet{numb}{magenta!60!black}

\lstdefinelanguage{json}{
    basicstyle=\scriptsize\ttfamily,
    showstringspaces=false,
    breaklines=true,
    frame=single,
    backgroundcolor=\color{white},
    literate=
     *{0}{{{\color{numb}0}}}{1}
      {1}{{{\color{numb}1}}}{1}
      {2}{{{\color{numb}2}}}{1}
      {3}{{{\color{numb}3}}}{1}
      {4}{{{\color{numb}4}}}{1}
      {5}{{{\color{numb}5}}}{1}
      {6}{{{\color{numb}6}}}{1}
      {7}{{{\color{numb}7}}}{1}
      {8}{{{\color{numb}8}}}{1}
      {9}{{{\color{numb}9}}}{1}
      {:}{{{\color{punct}{:}}}}{1}
      {,}{{{\color{punct}{,}}}}{1}
      {\{}{{{\color{delim}{\{}}}}{1}
      {\}}{{{\color{delim}{\}}}}}{1}
      {[}{{{\color{delim}{[}}}}{1}
      {]}{{{\color{delim}{]}}}}{1},
}

    \usepackage[preprint]{neurips_2025}

\usepackage[utf8]{inputenc} 
\usepackage[T1]{fontenc}    
\usepackage{hyperref}       
\usepackage{url}            
\usepackage{booktabs}       
\usepackage{amsfonts}       
\usepackage{nicefrac}       
\usepackage{microtype}      
\usepackage{xcolor}         

\usepackage{amsmath}

\usepackage{geometry}
\usepackage{inconsolata}
\newtcolorbox{promptbox}[1][]{
  enhanced,
  breakable,
  colback=gray!2,
  colframe=black!50,
  coltitle=white,
  colbacktitle=blue!60!black,
  fonttitle=\bfseries,
  title=#1,
  fontupper=\ttfamily\small,
  boxrule=0.6pt,
  arc=2mm,
  left=6pt,
  right=6pt,
  top=5pt,
  bottom=5pt,
  titlerule=1pt,
  separator sign none
}

\usepackage{listings}
\usepackage{algorithm2e}
\usepackage{multirow}
\usepackage{adjustbox}
\usepackage{xcolor}

\title{\sysname: A Text-to-SQL Agent with Self-Refinement, Consensus Enforcement, \\and Column Exploration}

\author{
    Minghang Deng$^{1}$ \quad Ashwin Ramachandran$^{1}$ \quad Canwen Xu$^{2}$  \\
    \textbf{\ Lanxiang Hu$^{1,2}$ \quad Zhewei Yao$^{2}$ \quad Anupam Datta$^{2}$ \quad Hao Zhang$^{1,2}$\thanks{Correspondence to haozhang@ucsd.edu}} \\
    $^1$University of California, San Diego \quad $^2$Snowflake AI Research
}

\newcommand{\sysname}{\text{ReFoRCE}}

\begin{document}

\maketitle

\input{text/0_abstract.tex}

\input{text/1_intro.tex}
\input{text/2_related.tex}
\input{text/3_methods.tex}
\input{text/4_experiments}
\input{text/5_conclusion.tex}


\newpage

\bibliography{reforce_nips25}
\bibliographystyle{plainnat}
\appendix
\input{text/6_appendix}

\end{document}

%% file: math_commands.tex
\usepackage{amsmath,amsfonts,bm}

\def\eqref#1{equation~\ref{#1}}

\def\1{\bm{1}}

\DeclareMathAlphabet{\mathsfit}{\encodingdefault}{\sfdefault}{m}{sl}
\SetMathAlphabet{\mathsfit}{bold}{\encodingdefault}{\sfdefault}{bx}{n}



%% file: text/0_abstract.tex
\begin{abstract}

We present ReFoRCE, a Text-to-SQL agent that tops the Spider 2.0 leaderboard—a challenging benchmark reflecting complex, real-world Text-to-SQL scenarios. While Text-to-SQL systems enable natural language queries over structured databases, deploying them in enterprise environments remains difficult due to large, complex schemas (with over 1,000 columns), diverse SQL dialects (e.g., BigQuery, Snowflake), and sophisticated query requirements (e.g., transformations and analytics). ReFoRCE addresses these challenges through: (a) \emph{database information compression} via pattern-based table grouping and LLM-guided schema linking to alleviate long-context issues; (b) \emph{self-refinement} to iteratively correct syntax and semantic errors across dialects; (c) \emph{majority-vote consensus} to select high-confidence candidates while deferring ambiguous cases arising from sophisticated queries; and (d) \emph{iterative column exploration} guided by execution feedback to resolve those deferred cases. ReFoRCE achieves new state-of-the-art results, with scores of 35.83 on Spider 2.0-Snow and 36.56 on Spider 2.0-Lite.\footnote{Code available on Github: \url{https://github.com/hao-ai-lab/ReFoRCE}.}


\end{abstract}

%% file: text/1_intro.tex
\section{Introduction}
\label{sec:introduction}
Text-to-SQL converts natural language into SQL queries, enabling intuitive access to relational databases~\citep{zelle1996learning,yu2018spider,lei2024spider}. It facilitates key applications like business intelligence and automation by reducing the burden on analysts and developers.

Despite strong performance on benchmarks like Spider 1.0~\citep{yu2018spider} and BIRD~\citep{li2024can}, where models surpass 90\% and 70\% execution accuracy respectively~\citep{talaei2024chess, pourreza2024chase, xie2025opensearch, li2025omnisql, pourreza2025reasoning}, prior Text-to-SQL methods operate in simplified, toy settings—with small schemas (50 columns/DB), a single SQL dialect (i.e. SQLite) with limited data type support, low SQL complexity (30 tokens/SQL), and minimal ambiguity~\citep{zhang2023act, gao2023text, pourreza2024din, li2025omnisql, pourreza2024chase}. These conditions fail to reflect the complexity of real-world scenarios, as evidenced by the sharp performance drop on the recently introduced Spider 2.0 benchmark~\citep{lei2024spider}, which involves large, cross-domain databases, multiple SQL dialects, diverse and nested data types, ambiguous queries, and complex reasoning tasks. This gap calls for more robust agentic pipelines that go beyond single-shot SQL generation—incorporating schema linking, value retrieval, candidate voting, and selection~\citep{talaei2024chess, pourreza2024chase, xie2025opensearch}—to meet the demands of real-world deployment.

Unlike general-purpose agents, Text-to-SQL agents face unique challenges in realistic settings: they must reason over large, complex schemas, handle nested or under-documented data types, and generate executable, dialect-specific SQL. These challenges are compounded by long-context constraints, where failures in schema tracking, instruction following, or ambiguity resolution can lead to execution errors or semantic drift. 

While recent methods have adopted tool-augmented reasoning and planning paradigms~\citep{wei2022chain,wang2023plan,shinn2024reflexion}, they are rarely adapted to the structured, high-stakes environment of Text-to-SQL. To bridge this gap, \cite{lei2024spider} proposed Spider-Agent, extending the ReAct paradigm~\citep{yao2023react} to incorporate schema navigation and step-by-step SQL construction. However, code agents often struggle to maintain control, especially in long-context scenarios where they may fail to follow instructions, overlook critical schema elements, or produce inconsistent and unreliable outputs~\citep{xia2024agentless}. In addition, existing Text-to-SQL methods~\citep{talaei2024chess, pourreza2024chase, xie2025opensearch} remain limited in their ability to handle multi-dialect SQL generation, deeply nested columns, and complex or under-specified data types, commonly seen in real databases.

\begin{figure}
    \centering
    \includegraphics[width=1\linewidth]{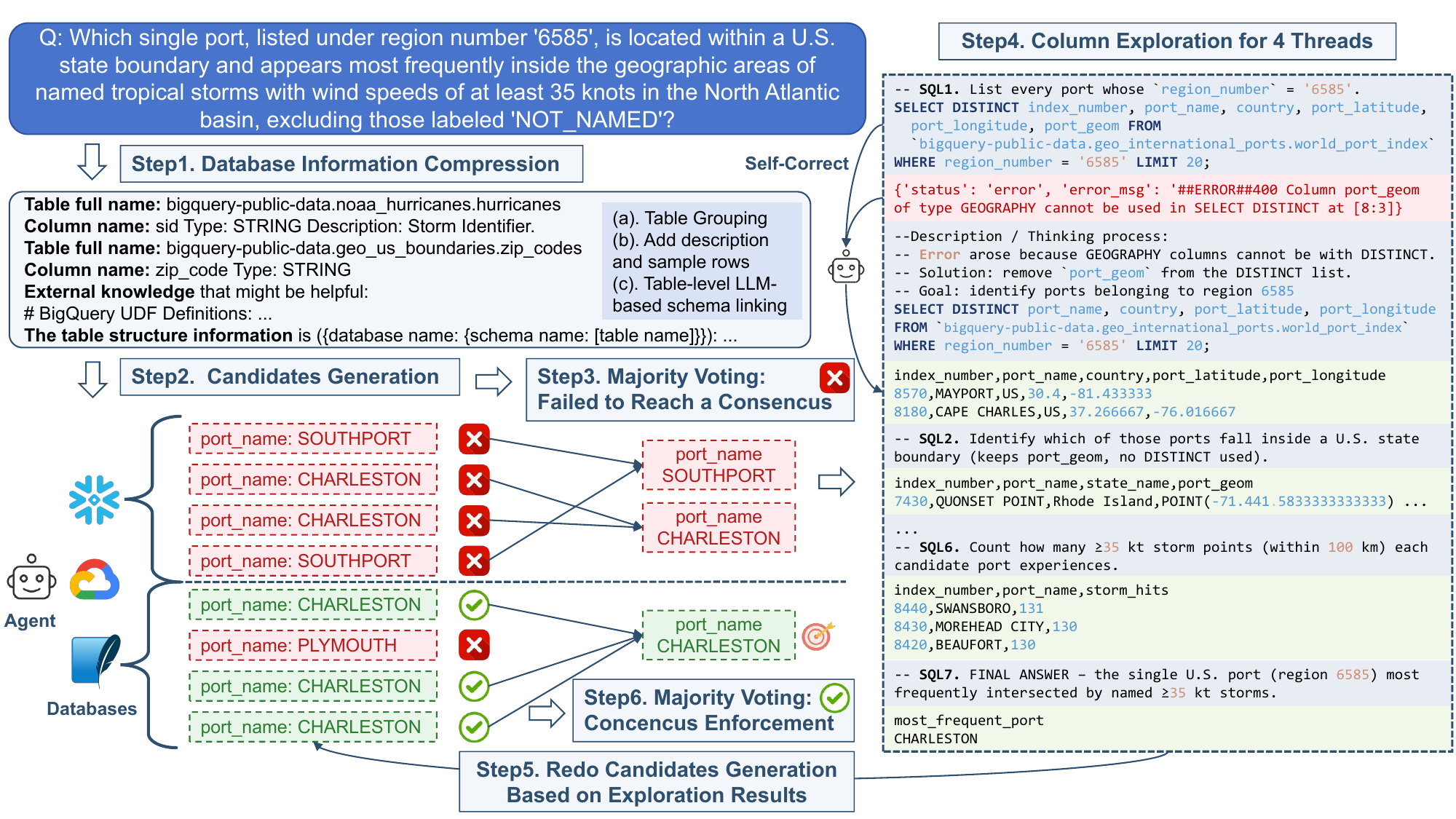}
    \caption{An overview of Self-\textbf{Re}finement Agent with Consensus En\textbf{FoR}cement and \textbf{C}olumn \textbf{E}xploration (\textbf{ReFoRCE}), which consists: (a) Database Information Compression, (b) Candidates Generation, (c) Majority Voting and (d) Column Exploration.}
    \label{fig:workflow}
\end{figure}

To address these challenges, we propose \textbf{ReFoRCE} (Self-\textbf{Re}finement Agent with Consensus En\textbf{FoR}cement and \textbf{C}olumn \textbf{E}xploration), a modular Text-to-SQL agent designed to decompose the task into controllable, purpose-driven subtasks. As illustrated in Figure~\ref{fig:workflow}, ReFoRCE begins with \emph{database information compression}, which tackles the long-context bottleneck by applying pattern-based table grouping and LLM-guided schema linking~\citep{talaei2024chess}---significantly reducing schema size while preserving task-relevant semantics, thereby simplifying reasoning over massive, heterogeneous schemas. Next, it performs \emph{candidate generation with self-refinement}, iteratively repairing and validating outputs to ensure dialect correctness and reduce semantic drift. To ensure consistency and confidence, ReFoRCE then applies \emph{majority-vote consensus}~\citep{pourreza2024chase, xie2025opensearch}, aggregating multiple candidates to filter out unreliable responses and defer ambiguous cases---offering robustness against instruction-following failures and hallucinated outputs. Finally, for the remained challenging examples, ReFoRCE introduces an optional \emph{column exploration} module, iteratively querying the database to resolve ambiguous references and handle nested or under-specified structures. Column exploration is selectively triggered, enabling the agent to balance accuracy and efficiency based on the difficulty of each example. By structuring the reasoning process into discrete, adaptable steps, ReFoRCE enhances control, reliability, and scalability in real-world Text-to-SQL generation.



ReFoRCE tops the Spider 2.0 leaderboard, achieving state-of-the-art performance on both the Snow and Lite subsets. It outperforms Spider-Agent~\citep{lei2024spider} baselines, scoring 35.83 on Snow and 36.56 on Lite. ReFoRCE is also efficient, requiring only 3.52 LLM and 3.89 DB calls per example. When column exploration is disabled, the cost drops to just 1.69 calls and 15K tokens per example, with minimal accuracy loss, offering a favorable balance between accuracy and efficiency. ReFoRCE further generalizes well to smaller models, boosting Qwen2.5-7B~\citep{yang2024qwen2} and Arctic-Text2SQL-7B~\citep{yao2025arctic} by over 10--20\% Execution Accuracy (EX) when integrated into our agentic workflow. Ablation studies confirm that database information compression is the most critical component, while column exploration significantly enhances EX@8 by promoting diverse candidate generation. Notably, using gold table annotations results in only marginal improvements, suggesting that strong reasoning models can infer relevant schema structures without explicit supervision in real-world Text-to-SQL scenarios~\citep{maamari2024death}.

In summary, our contributions are threefold: (1) We propose \textbf{ReFoRCE}, an agentic framework combining schema compression, self-refinement, consensus enforcement, and column exploration; (2) We establish new state-of-the-art results on Spider 2.0 while reducing resource overhead; and (3) We show ReFoRCE's compatibility with smaller open-source models, enabling competitive performance in low-resource settings.

%% file: text/2_related.tex
\section{Related Work}

\paragraph{Text-to-SQL Methods.} Recent advances in Text-to-SQL primarily fall into two categories: fine-tuning and large language model (LLM) prompting. Fine-tuning methods~\citep{wang2019rat, scholak2021picard, li2023resdsql, CodeS} focus on adapting models to specific benchmarks by learning schema representations, query structures, and logical relationships. In contrast, LLM prompting~\citep{zhang2023act, gao2023text, pourreza2024din, talaei2024chess} leverages carefully designed prompts in few-shot or zero-shot settings to avoid task-specific training. Although these methods perform well on simpler datasets like Spider 1.0~\citep{yu2018spider} and BIRD~\citep{li2024can}, they struggle with more complex benchmarks such as Spider 2.0~\citep{lei2024spider} due to challenges in schema understanding, ambiguity resolution, and dialect handling. Techniques such as semantic matching~\citep{kothyari-etal-2023-crush4sql}, in-context learning~\citep{gao2023texttosqlempoweredlargelanguage}, sub-query decomposition~\citep{pourreza2024din}, and memory-based self-refinement~\citep{shinn2024reflexion} have pushed the boundaries of SQL generation. Moreover, calibration methods like log probability scoring~\citep{ramachandran2024texttosqlcalibrationneedask} and binary validation~\citep{tian2023justaskcalibrationstrategies} have improved confidence estimation. Recently, reinforcement learning—especially Group Relative Policy Optimization (GRPO)\citep{shao2024deepseekmath}, which eliminates critic models in favor of rule-based rewards—has enabled the development of new SQL reasoning models\citep{li2025omnisql, pourreza2025reasoning} that surpass earlier supervised approaches.

\paragraph{Coding Agents.} Coding agents enable LLMs to interact dynamically with their environment by using tools, executing commands, observing feedback, and planning actions. Early frameworks like ReAct~\citep{yao2023react} introduced reasoning and acting components, while Self-Debugging~\citep{chen2023teaching} and InterCode~\citep{yang2024intercode} showcased iterative problem-solving through debugging and lightweight reinforcement learning. Plan-and-Solve Prompting~\citep{wang2023plan} and multi-agent systems like CodeR~\citep{chen2024coder} and Reflexion~\citep{shinn2024reflexion} further enhanced task decomposition and iterative improvement. Specialized frameworks, such as Spider Agent~\citep{lei2024spider}, addressed domain-specific challenges like SQL query generation. However, coding agents often face limitations in specialized tasks, where domain-specific solutions may outperform generalized frameworks~\citep{xia2024agentless}.

Another line of agent research target on structured and predefined workflows that guide LLMs and tools for more reliable performance designed for specific tasks. Originating from concepts like Chain-of-Thought~\citep{wei2022chain} and Self-Consistency~\citep{wang2022self}, workflows have evolved with advancements~\citep{josifoski2023flows,wu2023autogen,zeng2023flowmind}, enabling modular, collaborative, and automated workflows. In the Text-to-SQL setting, structured workflows decompose the overall task into subtasks such as question parsing, schema linking, value retrieval, candidate generation, and candidate selection~\citep{talaei2024chess,xie2025opensearch,pourreza2024chase}. Each step can be handled by individual or multiple agents, highlighting the effectiveness of modular workflows in complex coding tasks. While existing coding agents and workflows focus on iterative refinement and modular problem-solving, most Text-to-SQL agents are evaluated primarily on simpler datasets such as BIRD~\citep{li2024can} and Spider 1.0~\citep{yu2018spider}. In contrast, \textbf{ReFoRCE} introduces a unique combination of table compression, consensus enforcement, and iterative column exploration to address the challenges of enterprise-scale SQL generation.

%% file: text/3_methods.tex
\section{Methodology}
\label{sec:methodology}
\subsection{Preliminaries}
Spider 2.0~\citep{lei2024spider} is a comprehensive code agent task where, given a question $\mathcal{Q}$, a database interface $\mathcal{I}$, and a codebase $\mathcal{C}$ (including context, configuration, and documentation), the goal is to iteratively modify the code (SQL/Python) $\mathcal{C}$ based on observations $\mathcal{O}_k = \texttt{execute}(\mathcal{C}, \mathcal{I}, \mathcal{Q})$ until the final result $\mathcal{A}$ (text/table/database) is obtained. The final observation $\mathcal{O}_k$ serves as the agent’s answer to the question, i.e., $\mathcal{A} = \mathcal{O}_k$. In contrast, Spider 2.0-Snow and Spider 2.0-Lite are self-contained Text-to-SQL tasks. Given a database schema $\mathcal{D}$, a natural language question $\mathcal{Q}$, and auxiliary documentation $\mathcal{E}$, the Text-to-SQL parser $f(\cdot)$ generates the SQL query $\mathcal{S} = f(\mathcal{Q}, \mathcal{D}, \mathcal{E} \,|\, \theta)$, where $\theta$ denotes the parser’s parameters.

\cite{lei2024spider} introduced Spider-Agent, a framework built on the ReAct~\citep{yao2023react} paradigm with function-calling capabilities such as \texttt{EXEC\_SQL} for executing SQL queries and \texttt{TERMINAL} for performing command-line operations to navigate Data Build Tool (DBT) projects and read schema-related files. The agent operates by receiving observations, which represent the current state of the environment or the outcome of a function call initiated by the agent. Based on these observations, the agent generates a ``Thought'' and selects an appropriate ``Action'' from a predefined list of function calls. The task is considered complete when the agent invokes the \texttt{TERMINATE} function.

\subsection{\sysname: Self-Refinement, Consensus Enforcement, and Column Exploration}
ReAct-like agents, while flexible, often suffer from unreliable and unpredictable workflows due to their unconstrained decision-making process. To overcome these limitations, we propose \textbf{ReFoRCE} (Self-\textbf{Re}finement Agent with Consensus En\textbf{FoR}cement and \textbf{C}olumn \textbf{E}xploration), a structured framework that decomposes the Text-to-SQL task into controllable and reliable subtasks. As shown in Figure~\ref{fig:workflow}, ReFoRCE consists of (1) database information compression, where we reduce long-context input using pattern-based table grouping, LLM-based schema linking, and selective retention of column descriptions and sample rows; (2) candidate generation with self-refinement, where agents iteratively improve SQL predictions and eliminate syntax errors or empty results; (3) majority voting and result selection to enforce consensus among high-confidence outputs; and (4) iterative column exploration, which is triggered on low-confidence predictions to handle difficult examples involving nested schemas or complex task descriptions. This structured design ensures reliability in execution while maintaining flexibility in handling enterprise-scale SQL generation tasks.

\subsubsection{Database Information Compression}
\begin{figure}
    \centering
    \includegraphics[width=1\linewidth]{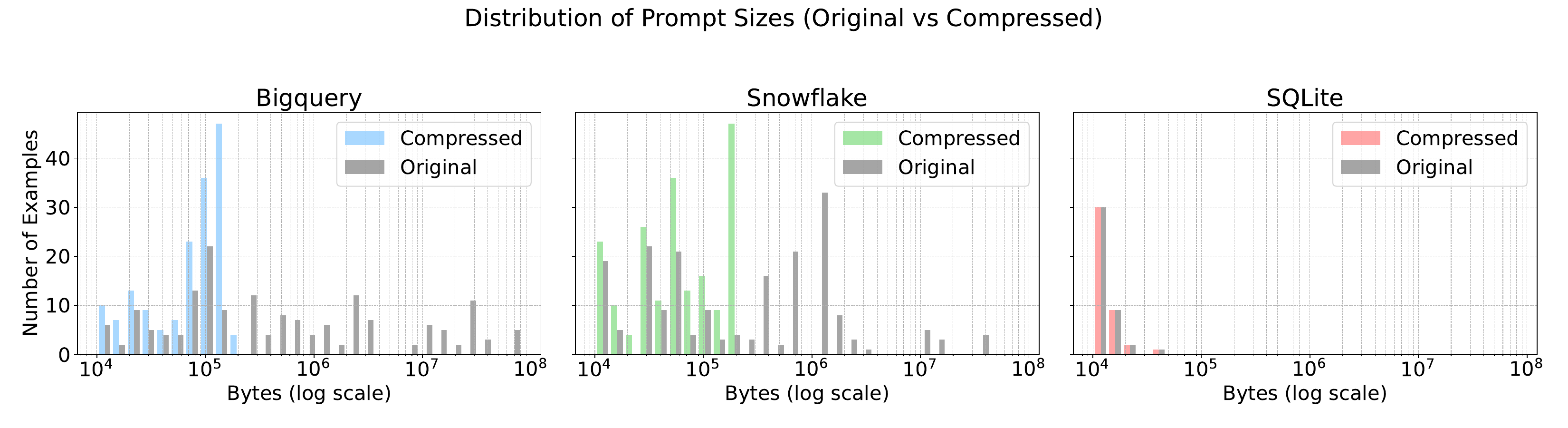}
    \caption{Distribution of prompt sizes before and after compression on the Spider 2.0-Lite dataset, grouped by dialects (BigQuery, Snowflake, and SQLite).}
    \label{fig:compression_comparison_plot}
\end{figure}

Previous work~\citep{talaei2024chess,pourreza2024chase,xie2025opensearch,pourreza2025reasoning} pays little attention to database information compression, primarily because datasets such as BIRD~\citep{li2024can} and Spider~\citep{yu2018spider} use lightweight local SQLite databases. These databases typically contain only a limited number of tables and columns, with a small variety of column types, making compression unnecessary—table information can be fully included in the prompt without exceeding the model’s context length. However, Spider 2.0 presents a different challenge: it includes databases with over 800 columns. If we incorporate sample rows and column descriptions as commonly done in BIRD, the schema length can reach up to 77MB (approximately 20 million tokens), as illustrated in Figure~\ref{fig:compression_comparison_plot}, which far exceeds the context length limits of current large language models (LLMs).

To mitigate this, we first apply a pattern-based matching strategy to merge tables with similar prefixes or suffixes. Among these, only one representative table is retained in full, while the rest are represented by their table names. Additionally, for examples still with long prompts (i.e., context length > 50K tokens) after table grouping, we adopt a simple, table-level LLM-based schema linking method inspired by~\cite{talaei2024chess}. This approach preserves reasoning space in the context while minimizing information (database schema, column description and sample rows) loss during linking. The prompt template and accuracy for schema linking is provided in Appendix~\ref{appendix:schema_linking}.

For example, in the \texttt{GA360} database, there is one year of data with table names ranging from \texttt{GA\_SESSIONS\_20160801} to \texttt{GA\_SESSIONS\_20170801}. Each table’s Data Definition Language (DDL) file occupies more than 150 KB, resulting in a total length of over 50 MB—an impractically large size for LLMs to process. 
Our pattern-based compression significantly reduces DDL file sizes of such databases to be less than 2 MB, more than 96\% compression without information loss.

Given the database information $\mathcal{D}$ and auxiliary documentation $\mathcal{E}$, we apply the \texttt{compress} function to compress it and concatenate the result with the question $\mathcal{Q}$ as the initial input prompt $\mathcal{P}_{\text{init}}$: 
\begin{equation}
    \mathcal{P}_{\text{init}} = \texttt{compress}(\mathcal{D}) + \mathcal{E} + \mathcal{Q}.
\end{equation}

\subsubsection{Generation and Consensus Enforcement}
Self-consistency and majority voting are commonly used in agentic Text-to-SQL workflows~\citep{talaei2024chess,pourreza2024chase,xie2025opensearch}, as they enhance both the \textit{stability} and \textit{accuracy} of the results—particularly for reasoning-intensive generation tasks, where models typically operate with a fixed temperature \( T = 1.0 \), resulting in diverse outputs. While many existing methods perform random selection from candidates after majority voting, this approach may introduce instability and fails to fully leverage the generated outputs. Instead of performing random selection, we retain ambiguous examples for further refinement.

For each input example \( x \), we generate \( k \) self-refined candidate results:
\begin{equation}
\mathcal{R}(x) = \{y_1, y_2, \dots, y_k\} = \texttt{LLM}(\mathcal{P}_{\text{init}}).
\end{equation}
We define the vote count for each candidate \( y_i \) as:
\begin{equation}
v(y_i) = \left| \{y_j \in \mathcal{R}(x) \mid y_j = y_i \} \right|.
\end{equation}
Let \( y^* = \arg\max_{y_i \in \mathcal{R}(x)} v(y_i) \) denote the candidate with the highest number of votes. If
\begin{equation}
\max_{y_i} v(y_i) > \max_{y_j \neq y_i} v(y_j),
\end{equation}
i.e., there is no tie, then \( y^* \) is considered a high-confidence answer and selected as the final result. In this manner, we transform diverse outputs into a consensus with high confidence.

The high-confidence example set \(\mathcal{X}_{\text{high}}\) is defined as:
\begin{equation}
\label{eq:high_conf}
\mathcal{X}_{\text{high}} = \left\{ x \in \mathcal{X} \,\middle|\, \exists! \, y^* \in \mathcal{R}(x) \text{ s.t. } v(y^*) = \max_{y \in \mathcal{R}(x)} v(y) \right\},
\end{equation}
where \(\exists!\) denotes the existence of a unique candidate with the highest vote count.

In contrast, if there exists a tie among candidates with the highest vote count, we define a low-confidence result set \(\mathcal{X}_{\text{low}}\) as:
\begin{equation}
\label{eq:low_conf}
\mathcal{X}_{\text{low}} = \left\{ x \in \mathcal{X} \,\middle|\, \exists\, y_i \neq y_j \text{ s.t. } v(y_i) = v(y_j) = \max_{y \in \mathcal{R}(x)} v(y) \right\}.
\end{equation}
These examples are retained for additional refinement steps to improve final answer selection.

\subsubsection{Column Exploration}
\label{subsec:CE}
When directly providing the entire database information to an LLM, the lack of details on value types and SQL dialects often leads to repeated iterations for refining syntax errors, correcting data types, or invoking the correct functions. This process is not only time-consuming but also leaves little room for real reasoning. For baselines such as DAIL-SQL~\citep{gao2023text} and DIN-SQL~\citep{pourreza2024din} in Spider 2.0-Lite, sample rows provided by the Lite dataset are often used to help the model understand the structure of the tables. However, for nested columns, even a few rows can be too lengthy to be fed into the LLM. Additionally, specific values in sample rows often lack diversity and are biased, which can mislead the model into generating incorrect answers based on these limited samples.
        
To address these challenges and ensure a comprehensive understanding of the database structure, we design a systematic approach for exploring potentially useful columns. The process begins by identifying relevant tables and columns through prompts tailored to extract meaningful information. Dynamically generated SQL queries then progress from simple, non-nested forms to increasingly complex structures, enabling a gradual and thorough understanding of the database to ultimately derive the correct answer. Appendix~\ref{appendix:column_exploration} presents the prompts and algorithm used for column exploration, while Appendix~\ref{appendix:case} illustrates a successful example. A key insight is that if an error occurs in one exploration query—such as misreading values from nested columns—we assume similar errors may occur in subsequent queries. We use LLMs to refine the faulty query and adjust related ones accordingly.

We apply column exploration to the unsolved set with low confidence defined in Equation~\ref{eq:low_conf}, and merge the resulting answers with the high-confidence set. Finally, we apply random selection to any remaining unresolved examples. Formally, let \texttt{column\_explore} denote the set of newly resolved examples from column exploration, and let \(\mathcal{X}_{\text{final}}\) denote the full set of resolved answers:
\begin{equation}
\mathcal{X}_{\text{final}} = \mathcal{X}_{\text{high}} \cup \texttt{column\_explore}(\mathcal{X}_{\text{low}}) \cup \texttt{random\_select}(\mathcal{X} \setminus \mathcal{X}_{\text{high}} \setminus \texttt{column\_explore}(\mathcal{X}_{\text{low}})),
\end{equation}
where \texttt{random\_select} denotes a final fallback strategy using random selection for the unresolved.

We do not apply column exploration at the beginning, not only due to its high computational cost but also because it often leads to more diverse outputs that slightly reduce accuracy, as shown in Table~\ref{tab:ablation_study_extended}. Column exploration is better suited for harder questions that demand broader search. For simpler questions, combining self-refinement with majority voting is typically sufficient to achieve strong results, while also being more cost-effective. This design balances exploration and exploitation, catering to both accuracy-driven scenarios and cost-efficient use cases.

%% file: text/4_experiments.tex
\section{Experiments}
\label{sec:experiments}
\subsection{Experimental Setup}

\paragraph{Dataset.} 
We evaluate our approach on the Spider 2.0 dataset~\citep{lei2024spider}, which comprises two Text-to-SQL tasks: Spider 2.0-Snow and Spider 2.0-Lite. Each task contains 547 examples drawn from over 150 databases, with an average of 800 columns per database. The SQL queries are complex, averaging around 150 tokens, posing a significant challenge. The main distinction between the tasks lies in their SQL dialects: Spider 2.0-Snow is limited to the Snowflake dialect, while Spider 2.0-Lite supports BigQuery, Snowflake, and SQLite. To evaluate smaller models trained on the SQLite dialect data within or without agentic workflow, we additionally consider the SQLite portion (135 examples) of Spider 2.0-Lite, referred to as Spider 2.0-SQLite. 

\paragraph{Evaluation Metrics.} We evaluate performance using the widely adopted metric \emph{Execution Accuracy (EX)}~\citep{yu2018spider, li2024can}. Compared to BIRD~\citep{li2024can}, Spider 2.0 employs a less strict evaluation script. In some cases, ambiguous questions may not clearly specify which columns to return, and the evaluation focuses on the essential components of the answers while ignoring irrelevant columns. As such, the inclusion of extra columns is deemed acceptable. Given that our method employs majority voting, we also report \textit{EX Pass@k} to reflect the best execution accuracy among the top-$k$ candidates, highlighting both prediction diversity and the upper performance bound. We estimate the average number of LLM and database API calls, as well as the tokens used per LLM call, to better assess the overall cost.

\paragraph{Large Language Models.} We conduct our experiments using models from the GPT family, including the non-reasoning model GPT-4o~\citep{achiam2023gpt} and the latest reasoning models o3 and o4-mini~\citep{openai2024api}. All models are evaluated with a fixed temperature of 1.0, except for greedy decoding setting of GPT-4o, whose temperature is 0.

\paragraph{Baselines.} For the main experiments on Spider 2.0-Snow and Spider 2.0-Lite, we use Spider-Agent~\citep{lei2024spider} with \{GPT-4o, o3, o4-mini\} as our baseline. For the Spider 2.0-SQLite subtask, we adopt Qwen2.5-7B-Instruct~\citep{yang2024qwen2} as the generation model baseline and OpenSearchSQL~\citep{xie2025opensearch} as the agentic workflow baseline. To achieve the best performance among 7B models on Spider 2.0-SQLite, we combine the current best-performing 7B Text-to-SQL generation model, Arctic-Text2SQL-R1-7B~\citep{yao2025arctic}, with ReFoRCE agentic workflow.

\subsection{Evaluation Results}

\begin{table}[t]
\centering
\caption{Comparison of methods on the Spider 2.0-Snow and Spider 2.0-Lite datasets. Metrics include Execution Accuracy (EX), EX Pass@8, per-example calls, and average number of tokens per LLM call. More details regarding costs estimation are available in Appendix~\ref{appendix:cost}. $^\dagger$: $v$ denotes number of votes. For reasoning models o3 and o4-mini on Spider-Agent, tokens used during the reasoning stage are excluded from the token count.}
\label{tab:method_comparison_extended}
\begin{adjustbox}{width=\linewidth}
\begin{tabular}{llccccccc}
\toprule
\multirow{2}{*}{\textbf{Method}} & \multirow{2}{*}{\textbf{Model}} 
& \multicolumn{2}{c}{\textbf{Snow}} & \multicolumn{2}{c}{\textbf{Lite}}
& \textbf{LLM Calls} & \textbf{DB Calls} 
& \textbf{Avg Tokens} \\
\cmidrule(r){3-4}
\cmidrule(l){5-6}
& & \textbf{EX} & \textbf{EX@8} 
& \textbf{EX} & \textbf{EX@8}  
& \textbf{/ example} & \textbf{/ example}
& \textbf{/ LLM call} \\
\midrule
\multirow{3}{*}{Spider-Agent~\citep{lei2024spider}} 
& GPT-4o   & 10.05 & - & 12.80 & - & \multirow{3}{*}{11$v$} & \multirow{3}{*}{3$v$}  & \multirow{3}{*}{8K$\cdot v^{\dagger}$} \\
& o4-mini  & 21.94 & - & 23.40 & - &  &  & \\
& o3       & 24.31 & - & 24.86 & - &  &  & \\
\midrule
\multirow{3}{*}{ReFoRCE} 
& GPT-4o   & 20.84 & 30.53 & 21.76 & 32.18 & \multirow{3}{*}{3.52$v$} & \multirow{3}{*}{3.89$v$} & 19K$\cdot v$ \\
& o4-mini  & 29.80 & 34.19 & 31.99 & 37.48 &  &  & 23K$\cdot v$ \\
& o3       & \textbf{35.83} & \textbf{39.85} & \textbf{36.56} & \textbf{42.05} &  &  & 23K$\cdot v$ \\
\bottomrule
\end{tabular}
\end{adjustbox}
\end{table}

\begin{table}[t]
\centering
\small
\caption{Comparison of performance on the Spider 2.0-SQLite subtask using smaller models as the generation component within an agentic workflow. Metrics include Execution Accuracy with Majority Voting (Maj., temperature = 1), Greedy Decoding (Greedy, temperature = 0), and EX Pass@8. $^\dagger$: Only the extraction components of OpenSearchSQL—schema linking and value retrieval—are used.}
\label{tab:method_tool_comparison}
\begin{tabular}{lccc}

\toprule
\multirow{2}{*}{\textbf{Method}} 
& \multicolumn{2}{c}{\textbf{Spider 2.0 - SQLite}}  \\
\cmidrule{2-3}
& \textbf{EX (Maj. / Greedy)} & \textbf{EX@8}\\
\midrule
ReFoRCE + GPT-4o & 24.44 / 22.22 & 37.03 \\
ReFoRCE + o4-mini & 31.11 / - & 39.26 \\
ReFoRCE + o3 & 38.52 / - & 42.22 \\
OpenSearchSQL$^\dagger$~\citep{xie2025opensearch} + ReFoRCE + o3 & 38.52 / - & 42.22 \\
\midrule
Qwen2.5-7B-Instruct~\citep{yang2024qwen2} & 5.19 / 4.44 & 7.40 \\
Arctic-Text2SQL-R1-7B~\citep{yao2025arctic} & 14.81 / 15.56 & 23.70 \\
\midrule
OpenSearchSQL + Qwen2.5-7B-Instruct & 7.40 / 4.44 & 10.37 \\
OpenSearchSQL + Arctic-Text2SQL-R1-7B & 14.07 / 17.78 & 20.74 \\
\midrule
ReFoRCE + o3 + Qwen2.5-7B-Instruct & 25.93 / 25.93 & 30.37 \\
\textbf{ReFoRCE + o3 + Arctic-Text2SQL-R1-7B} & 28.89 / \textbf{29.63} & \textbf{36.30} \\
\bottomrule

\end{tabular}
\end{table}

Table~\ref{tab:method_comparison_extended} compares the performance of different agentic frameworks on the Spider 2.0-Snow and Spider 2.0-Lite datasets. ReFoRCE demonstrates clear superiority over prior methods Spider-Agent~\citep{lei2024spider}, achieving the highest execution accuracy (EX) and EX@8 scores across both datasets. Specifically, ReFoRCE with the o3 model achieves 35.83\% and 36.56\% EX on Snow and Lite, respectively, with corresponding EX@8 scores of 39.85\% and 42.05\%. In comparison, Spider-Agent achieves performance that is roughly 10\% lower for each model.

In addition to achieving strong accuracy, ReFoRCE also maintains high efficiency when voting is not performed. It requires 3.52$v$ LLM calls and 3.89$v$ database calls per example, where $v$ denotes the number of voting candidates. Furthermore, if column exploration is disabled, the number of LLM and database calls per example can be further reduced to 1.69, with an average context length of approximately 15K tokens. This configuration yields a very efficient setup with only a minor performance drop of around 2\%, as shown in Table~\ref{tab:ablation_study_extended}, suitable for cost-efficient use cases. Additional details on cost estimation are provided in Appendix~\ref{appendix:cost}. In contrast, Spider-Agent makes approximately 11$v$ LLM calls per example and 8K$v$ tokens per call, which is significantly higher. Moreover, Spider-Agent runs inside a Docker container, which complicates concurrent execution, whereas ReFoRCE supports parallelism across examples, scaling up to the number of examples. In practice, the upper bound for concurrency is determined by the maximum number of concurrent API calls allowed by warehouses or LLMs, not by the ReFoRCE framework itself. These results highlight ReFoRCE’s strong balance between accuracy and efficiency, showcasing its capability to effectively combine model strength with structured agentic reasoning.

Table~\ref{tab:method_tool_comparison} presents a comparison of performance on the Spider 2.0-SQLite subtask using smaller models as the generation component within an agentic workflow. In this setup, the smaller models are responsible solely for SQL generation, while all other components are handled by the agentic workflow. The ReFoRCE framework achieves strong performance when combined with powerful proprietary models such as OpenAI's o3, reaching an EX score of 38.52 and an EX@8 score of 42.22. Incorporating OpenSearchSQL~\citep{xie2025opensearch}'s schema linking and value retrieval does not yield further gains, suggesting that ReFoRCE's internal mechanisms are already effective and that explicit schema linking and value retrieval may be unnecessary for this task. In contrast, when using smaller open-source models alone, performance declines significantly: Qwen2.5-7B-Instruct achieves only 5.19\% execution accuracy (Maj.), while Arctic-Text2SQL-R1-7B reaches 14.81\%. Adding the OpenSearchSQL extraction component yields minimal improvement for these models, indicating that schema linking and value retrieval are more beneficial for weaker models in this task. However, integrating these models with ReFoRCE and o3 substantially improves their effectiveness. For example, the combination of ReFoRCE + o3 + Arctic-Text2SQL-R1-7B~\citep{yao2025arctic} achieves a 29.63\% execution accuracy (Greedy) and 36.30\% EX@8, outperforming each individual component. This result surpasses GPT-4o and is comparable to o4-mini, which also underscores the complementary strengths of ReFoRCE’s structured reasoning and the SQL generation capabilities of smaller models, highlighting hybrid agentic workflows as a promising approach for enhancing performance with limited model capacity.

\subsection{Ablation Studies}
\label{subsec:ablation}
\begin{table}[t]
\centering
\caption{Ablation studies of the \textbf{ReFoRCE} framework (with o3) on Spider 2.0-Snow and Spider 2.0-Lite. $\Delta$EX and $\Delta$EX@8 denote the changes in execution accuracy and top-8 execution accuracy compared to the full model. $^\dagger$: ``Column Exploration Only'' refers to applying column exploration to all examples, rather than restricting it to the low-confidence set.}
\label{tab:ablation_study_extended}
\begin{adjustbox}{width=1\linewidth}
\begin{tabular}{lcccccccc}
\toprule
\multirow{2}{*}{\textbf{Method}} 
& \multicolumn{4}{c}{\textbf{Snow}} & \multicolumn{4}{c}{\textbf{Lite}} \\
\cmidrule(r){2-5}
\cmidrule(l){6-9}
& \textbf{EX} & \textbf{EX@8} 
& \textbf{$\Delta$EX} & \textbf{$\Delta$EX@8}
& \textbf{EX} & \textbf{EX@8} 
& \textbf{$\Delta$EX} & \textbf{$\Delta$EX@8} \\
\midrule
ReFoRCE (Full)                        & 35.83 & 39.85 & 0.00 & 0.00 & 36.56 & 42.05 & 0.00 & 0.00 \\
\midrule
ReFoRCE w/o Column Exploration        & 33.27 & 37.66 & -2.56 & -2.19 & 34.19 & 37.29 & -2.37 & \textbf{-4.76} \\
ReFoRCE w/o Random Selection          & 35.10 & 39.85 & -0.73 & -0.00 & 35.28 & 42.05 & -1.28 & -0.00 \\
ReFoRCE w/o Majority Voting           & 33.82 & 39.85 & -2.01 & -0.00 & 34.73 & 42.05 & -1.83 & -0.00 \\
ReFoRCE w/o DB Info Compression       & 32.18 & 36.20 & \textbf{-3.65} & \textbf{-3.65} & 33.09 & 38.57 & \textbf{-3.48} & -3.48 \\
ReFoRCE Column Exploration Only$^\dagger$ & 34.27 & 45.16 & -1.46 & +5.75 & 35.10 & \textbf{46.98} & -1.46 & \textbf{+4.93} \\
\midrule
ReFoRCE w/ Gold Table                  & \textbf{36.20} & \textbf{46.80} & \textbf{+0.37} & \textbf{+6.95} & \textbf{36.93} & \textbf{46.98} & \textbf{+0.37} & \textbf{+4.93} \\
\bottomrule
\end{tabular}
\end{adjustbox}
\end{table}

\begin{table}[t]
\centering
\caption{Impact of incorporating gold schema information into the \textbf{ReFoRCE} framework on Spider 2.0-Snow and Spider 2.0-Lite using gold SQLs. Note that both tasks provide only a subset of gold SQLs—121 examples for Snow and 256 for Lite—and their difficulty levels differ.}
\label{tab:gold_schema_ablation}
\begin{adjustbox}{width=1\linewidth}
\begin{tabular}{lcccccccc}
\toprule
\multirow{2}{*}{\textbf{Method}} 
& \multicolumn{4}{c}{\textbf{Snow}} & \multicolumn{4}{c}{\textbf{Lite}} \\
\cmidrule(r){2-5}
\cmidrule(l){6-9}
& \textbf{EX} & \textbf{EX@8} 
& \textbf{$\Delta$EX} & \textbf{$\Delta$EX@8}
& \textbf{EX} & \textbf{EX@8} 
& \textbf{$\Delta$EX} & \textbf{$\Delta$EX@8} \\
\midrule
ReFoRCE (Subset)            & 39.67 & 48.76 & 0.00 & 0.00 & 50.00 & 62.11 & 0.00 & 0.00 \\
ReFoRCE w/ Gold Schema     & 47.93 & 52.07 & \textbf{+8.26} & \textbf{+3.31} & 53.13 & 64.84 & \textbf{+3.13} & \textbf{+2.73} \\
\bottomrule
\end{tabular}
\end{adjustbox}
\end{table}

\paragraph{Column exploration promotes diverse generation.}
As shown in Table~\ref{tab:ablation_study_extended}, disabling column exploration results in a noticeable decrease in both EX and EX@8, particularly on the Lite subset (\textbf{-2.37} EX and \textbf{-4.76} EX@8), demonstrating its effectiveness. When column exploration is applied to all examples instead of only low-confidence ones, EX slightly decreases, while EX@8 shows a substantial increase (\textbf{+5.75} on Snow and \textbf{+4.93} on Lite), suggesting that broader exploration enhances diversity and benefits top-\emph{k} generation.

\paragraph{Database Information Compression contributes the most.}
Removing database information compression causes some examples to exceed the model’s maximum context length, resulting in failed executions and, consequently, no execution accuracy for those cases. This leads to the most significant performance drop, with \textbf{-3.65} EX and EX@8 on Snow and \textbf{-3.48} EX on Lite. As shown in Table~\ref{tab:ablation_study_extended}, this highlights the importance of schema compression in enabling large database inputs to fit within the model’s context window while retaining essential information for accurate reasoning.

\paragraph{Other design choices offer incremental improvements.}
Removing random selection or majority voting results in moderate performance drops (\textbf{-0.73} to \textbf{-2.01} EX), suggesting these strategies further stabilize performance, though they are less critical than schema compression or column exploration.

\paragraph{Gold tables offer limited gains on EX, but substantial improvement on EX@8.}
As shown in Table~\ref{tab:ablation_study_extended}, incorporating gold tables (as provided by Spider 2.0) leads to notable improvements in \textit{EX@8}—\textbf{+6.95} on Snow and \textbf{+4.93} on Lite—due to the significantly reduced schema search space. However, the overall gains in execution accuracy (\textit{EX}) are marginal, with only a \textbf{+0.37} increase on both subsets. This observation contrasts a little with assumptions in prior work on BIRD~\citep{pourreza2024chase,talaei2024chess,xie2025opensearch}, which emphasize gold schema selection as a critical factor. Several factors may contribute to this discrepancy:
(1) Table-level schema linking yields only modest benefits, whereas column-level schema linking and value retrieval remain essential~\citep{wang2025linkalign}, as further evidenced in Table~\ref{tab:gold_schema_ablation};
(2) Strong reasoning-capable models such as o3 can internally infer relevant schema information without explicit schema supervision, while schema linking continues to aid weaker models~\citep{maamari2024death}; and
(3) The Spider 2.0 benchmark emphasizes complex multi-step reasoning over pure retrieval, thereby reducing the marginal utility of gold table annotations.

\paragraph{Gold schema still matters.}
Incorporating gold schema information into the ReFoRCE framework consistently improves performance on both the Spider 2.0-Snow and Lite subsets. As shown in Table~\ref{tab:gold_schema_ablation}, ReFoRCE with gold schema achieves an absolute gain of \textbf{+8.26\%} in execution accuracy (EX) and +3.31\% in EX@8 on Snow. On the Lite subset, it improves by +3.13\% EX and +2.73\% EX@8. These results highlight that access to high-quality schema annotations continues to benefit Text-to-SQL agents—particularly in more challenging settings—even when the agent is capable of inferring schema from context. Notably, ReFoRCE is fully compatible with schema linking techniques, making it a strong foundation for future enhancements.

%% file: text/5_conclusion.tex
\section{Discussion}

\subsection{Limitations}
\label{sec:limits}
While ReFoRCE demonstrates strong performance on complex real-world datasets such as Spider 2.0, it exhibits limited generalization to simpler datasets like BIRD, which emphasize fine-grained schema-level understanding rather than long-context reasoning. Although ReFoRCE is compatible with external schema linking and value retrieval methods, we do not incorporate them in this work due to the lack of publicly available, high-quality implementations. Additionally, our approach relies on the internal reasoning capabilities of proprietary models such as o3 and does not involve any model training or fine-tuning of our own.

\subsection{Conclusion}
We introduced \textbf{ReFoRCE}, a structured agentic framework for Text-to-SQL generation over large, real-world databases. By decomposing the task into database compression, self-refinement, consensus enforcement, and column exploration, ReFoRCE improves controllability, accuracy, and scalability. It achieves state-of-the-art execution accuracy on the Spider 2.0 benchmark (Snow and Lite tasks) while maintaining efficiency in LLM and DB API usage. Extensive experiments highlight the contributions of each component, especially schema compression and column exploration for handling long contexts and complex queries. ReFoRCE also generalizes well to smaller open-source models, enabling strong performance under limited capacity.

Overall, ReFoRCE provides a modular and practical solution for real-world Text-to-SQL applications, enabling more reliable and efficient database querying in enterprise environments. Future work may explore tighter integration with external schema linking and retrieval modules, fine-tuning across multiple SQL dialects, and extending ReFoRCE to support broader analytical and transformation-oriented query tasks.

%% file: text/6_appendix.tex
\newpage
\appendix
\section{Table-level Schema Linking}
\label{appendix:schema_linking}

\begin{figure}
\begin{promptbox}
You are doing table level schema linking. Given a table with schema information and the task, you should think step by step and decide whether this table is related to the task. \\
You should answer Y/N only. If the answer is Y, you should add columns that you think related in python list format.\\
\\
Please answer only in json code block like:\\
\\
\{\\
    ``think'': ``think step by step to decide'',\\
    ``answer'': ``Y or N only'',\\
    ``columns'': [col\_name1, col\_name2]\\
\}\\
\\
\\
Table info: \{table\_info\}\\
Task: \{task\}\\
External knowledge: \{external\}
\end{promptbox}
\caption{Prompts for Table-level Schema Linking by Asking LLMs}
\label{prompt:schema_linking}
\end{figure}

Figure~\ref{prompt:schema_linking} illustrates prompts for table-level schema linking, where each table is queried individually using an LLM. The LLM is instructed to produce structured JSON output, including its reasoning process and a final answer restricted to ``Y'' or ``N'' only.

\begin{table}[t]
\centering
\caption{Recall and precision statistics After schema linking for Spider 2.0-Lite and Spider 2.0-Snow.}
\label{tab:recall_precision_stats}
\begin{tabular}{cccccc}
\toprule
\multirow{2}{*}{\textbf{Task}} & \multicolumn{3}{c}{\textbf{Long Context Examples}} & \multicolumn{2}{c}{\textbf{All Examples}} \\
\cmidrule(lr){2-4} \cmidrule(lr){5-6}
& \textbf{Recall} & \textbf{Precision} & \textbf{Recall = 1 Rate} 
& \textbf{Recall} & \textbf{Recall = 1 Rate} \\
\midrule
Lite & 86.24 & 40.81 & 48/63 (= 76.19\%) & 98.42 & 532/547 (= 97.26\%) \\
Snow & 85.32 & 30.24 & 50/57 (= 87.72\%) & 98.47 & 540/547 (= 98.72\%) \\
\bottomrule
\end{tabular}
\end{table}

Table~\ref{tab:recall_precision_stats} presents recall and precision statistics after the schema linking stage for both the Spider 2.0-Lite and Spider 2.0-Snow datasets. The results are reported separately for long-context subsets and for the complete set of examples. Schema linking is applied only to long-context cases (those exceeding 50K tokens), where our primary concern is recall. After schema linking, only 15 examples in Lite and 7 examples in Snow failed to achieve full recall, which we consider an acceptable trade-off given the overall coverage.

\section{API Cost}
\label{appendix:cost}

\begin{table}[t]
\centering
\caption{Statistics of Costs without Column Exploration: Distribution of self-refinement usage. $^*$: Maximum self-refinement iteration is capped at 5. $^\dagger$: For non-reasoning models.}
\label{tab:self_refinement_no_ce}
\begin{tabular}{lcccc}
\toprule
\textbf{Setting} & \textbf{Count} & \textbf{Avg LLM Calls} & \textbf{Avg DB Calls} & \textbf{Avg Tokens/LLM call$^\dagger$} \\
\midrule
No Self-refinement         & 407 & 1     & 1     & 15K \\
Little Self-refinement     & 90  & 3     & 3     & 16K \\
Empty Results              & 50  & 5$^*$ & 5     & 18K \\
\midrule
\textbf{Weighted Average}  & 547 & 1.69  & 1.69  & 15.44K \\
\bottomrule
\end{tabular}
\end{table}

\begin{table}[t]
\centering
\caption{Statistics of Costs with Column Exploration (CE) as the Second Step. $n$ denotes the number of voting candidates per example.}
\label{tab:ce_vs_no_ce}
\begin{adjustbox}{width=0.95\linewidth}
\begin{tabular}{lcccc}
\toprule
\textbf{Setting} & \textbf{Count} & \textbf{Avg LLM Calls} & \textbf{Avg DB Calls} & \textbf{Avg Tokens/LLM call} \\
\midrule
No Self-refinement \& No CE       & 407 & 1     & 1     & 15K \\
Little Self-refinement \& No CE   & 90  & 3     & 3     & 16K \\
Empty Results                     & 50  & 5     & 5     & 18K \\
With CE                           & 100 & 10    & 12     & 18K \\
\midrule
\textbf{Weighted Average}         & 547 & 3.52  & 3.89  & 18.73K \\
\textbf{Total Cost (×$v$ votes)}  & 547 & 3.52$v$ & 3.89$v$ & 18.73K$v$ \\
\bottomrule
\end{tabular}
\end{adjustbox}
\end{table}

In order to better understand the efficiency and cost-effectiveness of different strategies, we provide a detailed breakdown of LLM/database (DB) usage and token consumption under different refinement settings.

Table~\ref{tab:self_refinement_no_ce} reports the cost statistics without Column Exploration (CE), showing how many examples fall into different levels of self-refinement. Most examples (407) succeed without any self-refinement, consuming only a single LLM and DB call with a modest token count (15K). A smaller portion of examples require one or two iterations, while a few edge cases (50) reach the maximum of five iterations, incurring higher costs. The weighted average remains low at 1.69 LLM/DB calls and 15.44K tokens, demonstrating that self-refinement is generally efficient when CE is disabled.

Table~\ref{tab:ce_vs_no_ce} extends the analysis to include the Column Exploration (CE) step for cases that remain unresolved after self-refinement, which serves as the core strategy described in Section~\ref{subsec:CE}. While CE significantly increases cost—particularly for the 100 hard examples requiring CE, which incur on average 12 database calls and 30K tokens—the overall weighted average across all 647 examples remains acceptable, with 3.52 LLM calls, 3.89 DB calls, and 18.73K tokens per example. The last row of the table scales these averages by the number of candidates $v$ used in majority voting, reflecting the true per-example cost under the full ReFoRCE pipeline. This demonstrates the effectiveness of ReFoRCE’s two-stage design: inexpensive examples are handled efficiently via self-refinement alone, while costly CE operations are triggered only when needed, achieving a favorable balance between accuracy and efficiency.

\section{Column Exploration}
\label{appendix:column_exploration}

We design a prompt that instructs the LLM to generate up to 10 diverse \verb|SELECT| queries in the \verb|{api}| dialect, following the format of \verb|get_prompt_dialect_basic(api)|. Each query should explore different aspects of the data using only the provided tables (\verb|{table_struct}|), avoid schema inspection, and include \verb|LIMIT 20|. Annotations using \verb|--Description:| are required for each query. The related prompt is in Listing~\ref{lst:prompt_dialect} and Figure~\ref{prompt:column_exploration}.

Algorithm~\ref{alg:column_exploration_full} explores relevant database columns through a multi-step pipeline. It first generates candidate SQL queries from an initial prompt using an LLM. Each query is executed, and invalid ones are corrected via a self-correction loop with bounded retries. Valid results are collected, and if corrections are made, similar queries are refined to prevent repeated errors. A majority voting step identifies high-confidence answers. Finally, a column exploration prompt is constructed from the collected results and passed to the LLM to generate the final column-level exploration output $\mathcal{A}_{\text{explore}}$.
\begin{lstlisting}[caption={Prompt Template per SQL Dialect}, label={lst:prompt_dialect}]
def get_prompt_dialect_basic(api):
    if api == "snowflake":
        return SELECT "COLUMN_NAME" FROM DATABASE.SCHEMA.TABLE WHERE ... (Adjust "DATABASE", "SCHEMA", and "TABLE" to match actual names, ensure all column names are enclosed in double quotations)
    elif api == "bigquery":
        return SELECT `column_name` FROM `database.schema.table` WHERE ... (Replace `database`, `schema`, and `table` with actual names. Enclose column names and table identifiers with backticks.)
    elif api == "sqlite":
        return SELECT DISTINCT "column_name" FROM "table_name" WHERE ... (Replace "table_name" with the actual table name. Enclose table and column names with double quotations if they contain special characters or match reserved keywords.)

def get_prompt_dialect_nested(api):
    if api == "snowflake":
        return For columns in json nested format: e.g. SELECT t."column_name", f.value::VARIANT:"key_name"::STRING AS "abstract_text" FROM PATENTS.PATENTS.PUBLICATIONS t, LATERAL FLATTEN(input => t."json_column_name") f; DO NOT directly answer the task and ensure all column names are enclosed in double quotations. For nested columns like event_params, when you do not know the structure of it, first watch the whole column: SELECT f.value FROM table, LATERAL FLATTEN(input => t."event_params") f;
    elif api == "bigquery":
        return Extract a specific key from a nested JSON column: SELECT t."column_name", JSON_EXTRACT_SCALAR(f.value, "$.key_name") AS "abstract_text" FROM `database.schema.table` AS t, UNNEST(JSON_EXTRACT_ARRAY(t."json_column_name")) AS f;\nWhen the structure of the nested column (e.g., event_params) is unknown, first inspect the whole column: SELECT f.value FROM `project.dataset.table` AS t, UNNEST(JSON_EXTRACT_ARRAY(t."event_params")) AS f;

def get_prompt_dialect_string_matching(api):
    if api == "snowflake":
        return Do not directly match strings if you are not convinced. Use fuzzy query first: WHERE str ILIKE "%target_str%" For string matching, e.g. meat lovers, you should use % to replace space. e.g. ILKIE %meat%lovers%.
    elif api == "bigquery":
        return Do not directly match strings if you are not convinced. Use LOWER for fuzzy queries: WHERE LOWER(str) LIKE LOWER('%target_str%'). For example, to match 'meat lovers', use LOWER(str) LIKE '%meat%lovers%'.
    elif api == "sqlite":
        return Do not directly match strings if you are not convinced. For fuzzy queries, use: WHERE str LIKE '%target_str%'. For example, to match 'meat lovers', use WHERE str LIKE '%meat%lovers%'. If case sensitivity is needed, add COLLATE BINARY: WHERE str LIKE '%target_str%' COLLATE BINARY.

\end{lstlisting}

\begin{algorithm}[ht]
\caption{Column Exploration with LLM Self-Correction}
\label{alg:column_exploration_full}
\KwIn{Initial prompt $\mathcal{P}_{\text{init}}$}
\KwOut{Column Exploration Results $\mathcal{A}_{\text{explore}}$}

\SetKwFunction{LLM}{LLM}
\SetKwFunction{Vote}{Vote}
\SetKwFunction{ColumnExplore}{ColumnExploration}
\SetKwFunction{RandomSelect}{RandomSelect}
\SetKwFunction{Execute}{execute\_sql\_sqlite}
\SetKwFunction{Correct}{self\_correct}
\SetKwFunction{Refine}{similar\_error\_refine}

\tcp{Step 1: Generate SQLs from initial prompt}
\texttt{sqls} $\leftarrow$ \LLM{$\mathcal{P}_{\text{init}}$} \;
\texttt{result\_dic\_list} $\leftarrow \emptyset$ \;

\tcp{Step 2: Execute SQLs with self-correction loop}
\While{\texttt{sqls} not empty}{
    \texttt{sql} = \texttt{sqls[0]}\;
    \texttt{results} = \Execute{\texttt{sql}}\;

    \If{\texttt{results} is valid (i.e., string and not empty)}{
        Append \texttt{sql} and \texttt{results} to \texttt{result\_dic\_list}\;
        Continue to next SQL\;
    }

    Initialize \texttt{max\_try}\;
    \While{\texttt{results} is not valid}{
        \If{\texttt{max\_try == 0}}{
            \textbf{break}\;
        }
        \texttt{corrected\_sql} $\gets$ \Correct{\texttt{sql}, \texttt{results}}\;
        \If{\texttt{corrected\_sql} is not valid}{
            \textbf{continue}\;
        }
        \texttt{results} = \Execute{\texttt{corrected\_sql}}\;
        Decrease \texttt{max\_try} by 1\;
    }

    \If{\texttt{results} is valid}{
        \If{\texttt{sqls} not empty}{
            \texttt{sqls} $\gets$ \Refine{\texttt{sqls}}\;
        }
        \If{\texttt{corrected\_sql} exists}{
            Append \texttt{corrected\_sql} and \texttt{results} to \texttt{result\_dic\_list}\;
        }
    }
}

\tcp{Step 3: Majority voting}
$\mathcal{X}_{\text{high}} \leftarrow \emptyset$, $\mathcal{X}_{\text{low}} \leftarrow \emptyset$ \;
\ForEach{$x \in \mathcal{X}$}{
  compute vote counts $v(y_i)$ for each $y_i \in \mathcal{R}(x)$ \;
  \eIf{a unique $y^*$ with highest $v(y^*)$ exists}{
    $\mathcal{X}_{\text{high}} \leftarrow \mathcal{X}_{\text{high}} \cup \{x\}$ with answer $y^*$ \;
  }{
    $\mathcal{X}_{\text{low}} \leftarrow \mathcal{X}_{\text{low}} \cup \{x\}$ \;
  }
}

\tcp{Step 4: Column exploration prompt and inference}
construct $\mathcal{P}_{\text{column\_exploration}}$ from \texttt{result\_dic\_list} \;
$\mathcal{A}_{\text{explore}} \leftarrow$ \LLM{$\mathcal{P}_{\text{column\_exploration}}$} \;

\Return{$\mathcal{A}_{\text{explore}}$}
\end{algorithm}

\begin{figure}
\begin{promptbox}
Write at most 10 \{api\} SQL queries for simple to complex ones to final answer in format like:\\
\\
\{\texttt{get\_prompt\_dialect\_basic}(api)\}\\
\\
in sql code block to have an understanding of values in related columns.\\
\\
Each query should be different. Do not query about any SCHEMA or checking data types. You can write SELECT query only. Try to use DISTINCT. For each SQL LIMIT 20 rows.\\
\\
Write annotations to describe each SQL, format like\\
--Description: \\
SELECT ...\\
\\
\{\texttt{get\_prompt\_dialect\_nested}(api)\}\\
\\
\{\texttt{get\_prompt\_dialect\_string\_matching}(api)\}\\
\\
For time-related queries, given the variety of formats, avoid using time converting functions unless you are certain of the specific format being used.\\
\\
You can only use tables in \{table\_struct\}\\
Your knowledge is based on information in database. Do not use your own knowledge.
\end{promptbox}
\caption{Prompts for Column Exploration}
\label{prompt:column_exploration}
\end{figure}

\section{A Case Study for Column Exploration}
\label{appendix:case}

\paragraph{Step 1: Failure Without Column Exploration.}
As shown in Listing~\ref{lst:col_ex1}, the LLM is tasked with answering a complex geospatial SQL question: counting the number of overlapping road pairs in California that do not share any nodes and are not tagged as bridges. In this initial step, the model attempts to directly solve the task using only the given question and compressed database schema, without engaging in column-level exploration. It constructs a multi-step SQL query that joins a large set of OSM "ways" based on geometric intersections and node disjointness. However, despite several rounds of self-refinement, the model's queries consistently fail to execute due to dialect-specific limitations in Snowflake, such as disallowed lateral table functions in outer joins and nested subqueries that violate evaluation constraints. These repeated failures highlight a key limitation of relying solely on prompt-based generation for complex tasks: the model lacks sufficient visibility into the schema and tagging structure of the underlying data, resulting in syntactic and semantic errors that cannot be corrected without external feedback. In particular, the model misuses constructs like \texttt{TABLE(FLATTEN(...))} and struggles to reason about the appropriate usage of \texttt{EXISTS} versus \texttt{JOIN} in Snowflake’s SQL dialect. As a result, the task remains unsolved within the self-refinement loop, demonstrating the ineffectiveness of schema-agnostic generation on tasks requiring deep schema understanding and dialect compliance.

\paragraph{Step 2: Success With Column Exploration.}
Listing~\ref{lst:col_ex} presents the same task, but now augmented with ReFoRCE's column exploration mechanism. Rather than attempting to directly generate a full query, the model first issues a series of exploratory SQL commands designed to understand the structure of the \texttt{PLANET\_WAYS} table and its nested \texttt{all\_tags} field. These include inspecting a sample of tagged ways, identifying which \texttt{highway} values are present, verifying how bridge attributes are encoded, and confirming the availability of geometric and node-based metadata. By iteratively executing and analyzing the results of these subqueries, the model is able to form a grounded understanding of how road classification, spatial filtering, and bridge exclusion can be expressed through dialect-compliant SQL. The insights obtained from exploration are then incorporated into a robust final query that avoids previously failed patterns (e.g., nested \texttt{FLATTEN} within subqueries), instead using aggregation and conditional logic such as \texttt{MAX(IFF(...))} in \texttt{HAVING} clauses to express filtering over tag structures. This process culminates in a successful and efficient query that counts all overlapping road pairs in California without shared nodes or bridge tags, yielding a final answer of \texttt{13} pairs. The transition from failure to success across Listings~\ref{lst:col_ex1} and~\ref{lst:col_ex} underscores the critical importance of column exploration for enabling grounded, dialect-aware reasoning over complex database schemas. Rather than relying on a single-shot generation approach, the model incrementally builds up semantic and structural understanding of the data, allowing it to craft accurate and executable SQL in challenging real-world scenarios.

\begin{lstlisting}[caption={A Successful Case for Column Exploration: Step 1. No Column Exploration}, label={lst:col_ex1}]
Step 1. No Column Exploration:

Task: How many different pairs of roads classified as motorway, trunk, primary, secondary, or residential in California overlap each other without sharing nodes and do not have a bridge tag, where these roads are tagged with highway, analyzed through planet ways?

[Try to run SQL in self-refine]
WITH state_ca AS (     -- California boundary
    SELECT "state_geom" AS geom
    FROM GEO_OPENSTREETMAP_BOUNDARIES.GEO_US_BOUNDARIES.STATES
    WHERE "state" = 'CA'
),                                                        -- OSM roads that satisfy all filters
roads_ca AS (
    SELECT
        w."id"                                AS way_id ,
        w."geometry"                          AS geom_bin ,
        w."nodes"                             AS nodes_var
    FROM GEO_OPENSTREETMAP_BOUNDARIES.GEO_OPENSTREETMAP.PLANET_WAYS AS w ,
         state_ca AS s
    WHERE
          ST_INTERSECTS ( TO_GEOMETRY ( w."geometry" )
                         , TO_GEOMETRY ( s.geom ) )                -- lies in CA
      AND EXISTS (                                                  -- wanted highway class
              SELECT 1
              FROM  TABLE( FLATTEN ( INPUT => w."all_tags" ) ) AS t
              WHERE  LOWER( t.value:"key" )   = 'highway'
                AND  LOWER( t.value:"value") IN ( 'motorway'
                                                ,'trunk'
                                                ,'primary'
                                                ,'secondary'
                                                ,'residential' )
      )
      AND NOT EXISTS (                                              -- drop bridges
              SELECT 1
              FROM  TABLE( FLATTEN ( INPUT => w."all_tags" ) ) AS t
              WHERE  LOWER( t.value:"key") = 'bridge'
      )
)
SELECT
    COUNT(*) AS overlapping_road_pairs_without_shared_nodes
FROM   roads_ca  AS a
JOIN   roads_ca  AS b
       ON  a.way_id < b.way_id                                      -- unique unordered pairs
       AND ST_INTERSECTS( TO_GEOMETRY( a.geom_bin )
                        , TO_GEOMETRY( b.geom_bin ) )                -- geometries overlap
       AND NOT EXISTS (                                              -- but share no common node
              SELECT 1
              FROM  TABLE( FLATTEN ( INPUT => a.nodes_var ) ) AS n1
              JOIN  TABLE( FLATTEN ( INPUT => b.nodes_var ) ) AS n2
                    ON n1.value:"id" = n2.value:"id"
       );
[Try to run SQL in self-refine]
[Results]
{'status': 'error', 'error_msg': "##ERROR##000002 (0A000): Unsupported feature 'lateral table function called with OUTER JOIN syntax or a join predicate (ON clause)'."}
[Results]
[Try to run corrected SQL in self-refine]
WITH state_ca AS (   -- California boundary
    SELECT "state_geom" AS geom
    FROM GEO_OPENSTREETMAP_BOUNDARIES.GEO_US_BOUNDARIES.STATES
    WHERE "state" = 'CA'
),                   
roads_ca AS (        -- CA ways : wanted highway classes, not bridges
    SELECT
        w."id"        AS way_id,
        w."geometry"  AS geom_bin,
        w."nodes"     AS nodes_var
    FROM GEO_OPENSTREETMAP_BOUNDARIES.GEO_OPENSTREETMAP.PLANET_WAYS w,
         state_ca s
    WHERE ST_INTERSECTS(
              TO_GEOMETRY(w."geometry"),
              TO_GEOMETRY(s.geom)
          )
      AND EXISTS (                           -- motorway / trunk / primary / secondary / residential
            SELECT 1
            FROM TABLE(FLATTEN(INPUT => w."all_tags")) t
            WHERE  LOWER(t.value:"key")  = 'highway'
               AND LOWER(t.value:"value") IN
                    ('motorway','trunk','primary','secondary','residential')
      )
      AND NOT EXISTS (                       -- exclude bridges
            SELECT 1
            FROM TABLE(FLATTEN(INPUT => w."all_tags")) t
            WHERE LOWER(t.value:"key") = 'bridge'
      )
),
road_nodes AS (      -- one row per (way , node_id)
    SELECT
        r.way_id,
        f.value:"id"::NUMBER AS node_id
    FROM roads_ca r,
         LATERAL TABLE(FLATTEN(INPUT => r.nodes_var)) f
)

SELECT
    COUNT(*) AS overlapping_road_pairs_without_shared_nodes
FROM   roads_ca a
JOIN   roads_ca b
       ON  a.way_id < b.way_id
       AND ST_INTERSECTS(
               TO_GEOMETRY(a.geom_bin),
               TO_GEOMETRY(b.geom_bin)
           )                         -- geometries overlap
WHERE NOT EXISTS (                   -- but share no common node
          SELECT 1
          FROM   road_nodes rn1
          JOIN   road_nodes rn2
                 ON rn1.node_id = rn2.node_id
          WHERE  rn1.way_id = a.way_id
            AND  rn2.way_id = b.way_id
      );
[Try to run corrected SQL in self-refine]
[Results]
{'status': 'error', 'error_msg': "##ERROR##001003 (42000): SQL compilation error:\nsyntax error line 35 at position 17 unexpected 'TABLE'.\nsyntax error line 35 at position 22 unexpected '('."}
[Results]
[Try to run corrected SQL in self-refine]
WITH state_ca AS (   ----------------------------------------------------------------
    SELECT "state_geom" AS geom
    FROM GEO_OPENSTREETMAP_BOUNDARIES.GEO_US_BOUNDARIES.STATES
    WHERE "state" = 'CA'
),                   
roads_ca AS (        ----------------------------------------------------------------
    SELECT
        w."id"        AS way_id ,
        w."geometry"  AS geom_bin ,
        w."nodes"     AS nodes_var
    FROM GEO_OPENSTREETMAP_BOUNDARIES.GEO_OPENSTREETMAP.PLANET_WAYS w ,
         state_ca s
    WHERE  ST_INTERSECTS( TO_GEOMETRY(w."geometry")
                         , TO_GEOMETRY(s.geom) )
      AND EXISTS (      -- keep requested highway classes
              SELECT 1
              FROM   LATERAL FLATTEN( INPUT => w."all_tags" ) t
              WHERE  LOWER( t.value:"key" ) = 'highway'
                 AND LOWER( t.value:"value") IN
                     ('motorway','trunk','primary','secondary','residential')
      )
      AND NOT EXISTS (  -- exclude bridges
              SELECT 1
              FROM   LATERAL FLATTEN( INPUT => w."all_tags" ) t
              WHERE  LOWER( t.value:"key") = 'bridge'
      )
),
road_nodes AS (      ----------------------------------------------------------------
    SELECT
        r.way_id,
        n.value:"id"::NUMBER AS node_id
    FROM roads_ca r,
         LATERAL FLATTEN( INPUT => r.nodes_var ) n
)

SELECT
    COUNT(*) AS overlapping_road_pairs_without_shared_nodes
FROM   roads_ca a
JOIN   roads_ca b
       ON  a.way_id < b.way_id
       AND ST_INTERSECTS(
               TO_GEOMETRY( a.geom_bin ),
               TO_GEOMETRY( b.geom_bin )
           )                     -- geometries overlap
WHERE  NOT EXISTS (              -- but share no common node
           SELECT 1
           FROM   road_nodes rn1
           JOIN   road_nodes rn2
                  ON rn1.node_id = rn2.node_id
           WHERE  rn1.way_id = a.way_id
             AND  rn2.way_id = b.way_id
       );
[Try to run corrected SQL in self-refine]
[Results]
{'status': 'error', 'error_msg': '##ERROR##002031 (42601): SQL compilation error:\nUnsupported subquery type cannot be evaluated'}
[Results]

Max iteration reached, break.
\end{lstlisting}

\begin{lstlisting}[caption={A Successful Case for Column Exploration: Step 2. With Column Exploration}, label={lst:col_ex}]
Step 2. With Column Exploration:

[Try to execute]
-- Description: 1) Peek at a few OSM "ways" that have at least one tag whose key is 'highway'.
SELECT  w."id",
        w."all_tags"
FROM    GEO_OPENSTREETMAP_BOUNDARIES.GEO_OPENSTREETMAP."PLANET_WAYS"   w,
        LATERAL FLATTEN(input => w."all_tags")                         tag
WHERE   tag.value:"key"::STRING  ILIKE  '%highway%'
LIMIT 20;
[Try to execute]
[Successfully executed]
id,all_tags
386266471,"[
  {
    ""key"": ""crossing"",
    ""value"": ""uncontrolled""
  },
  {
    ""key"": ""footway"",
    ""value"": ""crossing""
  },
  {
    ""key"": ""highway"",
    ""value"": ""footway""
  }
]"
386266470,"[
  {
    ""key"": ""crossing"",
    ""value"": ""uncontrolled""
  },
  {
    ""key"": ""footway"",
    ""value"": ""crossing""
  },
  {
    ""key"": ""highway"",
    ""value"": ""footway""
  }
]"

[Successfully executed]
[Try to execute]
-- Description: 2) List the different "highway" values that appear in PLANET_WAYS (quick look-up).
SELECT DISTINCT 
        tag.value:"value"::STRING                             AS "highway_value"
FROM    GEO_OPENSTREETMAP_BOUNDARIES.GEO_OPENSTREETMAP."PLANET_WAYS" w,
        LATERAL FLATTEN(input => w."all_tags")                         tag
WHERE   tag.value:"key"::STRING ILIKE '%highway%'
LIMIT 20;
[Try to execute]
[Successfully executed]
highway_value
rest_area
path
motorway_link
primary
trunk
unclassified
footway
services
raceway
tertiary_link
yes
trunk_link
service
secondary
cycleway
track
bridleway
residential
motorway
pedestrian
[Successfully executed]
[Try to execute]
-- Description: 3) Inspect how the bridge information is carried in the tags of a sample way.
SELECT  w."id",
        tag.value:"key"::STRING     AS "tag_key",
        tag.value:"value"::STRING   AS "tag_value"
FROM    GEO_OPENSTREETMAP_BOUNDARIES.GEO_OPENSTREETMAP."PLANET_WAYS" w,
        LATERAL FLATTEN(input => w."all_tags") tag
WHERE   w."id" = (SELECT MIN("id") 
                  FROM GEO_OPENSTREETMAP_BOUNDARIES.GEO_OPENSTREETMAP."PLANET_WAYS")
LIMIT 20;
[Try to execute]
[Successfully executed]
id,tag_key,tag_value
4366743,bicycle,no
4366743,destination,Rest Area
4366743,highway,motorway_link
4366743,lanes,1
4366743,oneway,yes
[Successfully executed]
[Try to execute]
-- Description: 4) Pull out every California state geometry (there is only one row, but LIMIT just in case).
SELECT  "state_fips_code",
        "state_name",
        "state_geom"
FROM    GEO_OPENSTREETMAP_BOUNDARIES.GEO_US_BOUNDARIES."STATES"
WHERE   "state" = 'CA'
LIMIT 20;
[Try to execute]
[Successfully executed]
state_fips_code,state_name,state_geom
06,California,"bytearray(b'\x01\x06\x00\x00\x00\x07\x00\x00\x00\x01\x03\x00\x00\x00\x01\x00\x00\x00\x88\x01\x00\x00\xcf\x1b\xb5\xc2\xf4\xdd]\xc0\x12\rR\xf0\x14\x08A@\x99\x87\xda6\x8c\xde]\xc0R\x80(\x981\tA@\xb2\xf6w\xb6G\xdf]\xc0\xdc\x80\xcf\x0f#\nA@\x8b\x05L\xe0\xd6\xdf]\xc0LJA\xb7\x97\nA@""q\x8f\xa5\x0f\xe0]\xc0\xf4\xd3\x7f\xd6\xfc\nA@w\x85>X\xc6\xe0]\xc0\xf8\xbdM\x7f\xf6\x0bA@\xbf\xa41ZG\xe1]\xc0r1\x06\xd6q\x0cA@\xe9\xf7\x8b\xd9\x92\xe1]\xc0\x11i\x1b\x7f"
[Successfully executed]
[Try to execute]
-- Description: 5) Return a handful of OSM ways whose geometry intersects California 
--                and whose highway type is motorway|trunk|primary|secondary|residential.
SELECT  w."id",
        tag.value:"value"::STRING           AS "highway",
        w."geometry"
FROM    GEO_OPENSTREETMAP_BOUNDARIES.GEO_OPENSTREETMAP."PLANET_WAYS" w,
        GEO_OPENSTREETMAP_BOUNDARIES.GEO_US_BOUNDARIES."STATES"      s,
        LATERAL FLATTEN(input => w."all_tags")                       tag
WHERE   s."state" = 'CA'
  AND   tag.value:"key"::STRING  ILIKE  '%highway%'
  AND   tag.value:"value"::STRING ILIKE ANY ('%motorway%','%trunk%','%primary%','%secondary%','%residential%')
  AND   ST_INTERSECTS( TO_GEOGRAPHY(w."geometry"), TO_GEOGRAPHY(s."state_geom") )
LIMIT 20;
[Try to execute]
[Successfully executed]
id,highway,geometry
173040551,primary,"bytearray(b'\x01\x02\x00\x00\x00\x06\x00\x00\x000\xf9\x9f\xfc]\xcf]\xc0Mx\xae\x94\n\xcdA@\xe0\xa3Y4x\xcf]\xc0\x01\x7fB\x87\n\xcdA@\xbe\xa9\xa3>$\xd0]\xc0\xc6\xf6\xff\x16\t\xcdA@\x17\tm9\x97\xd0]\xc0\x91+\xf5,\x08\xcdA@\x9a[!\xac\xc6\xd0]\xc0\xcc]K\xc8\x07\xcdA@\xf9\xe1 !\xca\xd0]\xc0U\xe51\xb9\x07\xcdA@')"
173040548,primary,"bytearray(b""\x01\x02\x00\x00\x00\r\x00\x00\x00\x05\x7f\xael\xd5\xd0]\xc0U\xe51\xb9\x07\xcdA@ \xd7<\xfd\xd6\xd0]\xc0\xd8\x95;\xd8\x0"
[Successfully executed]
[Try to execute]
-- Description: 6) Same as above, but be sure the way has NO tag whose key = 'bridge'.
SELECT  DISTINCT w."id"
FROM    GEO_OPENSTREETMAP_BOUNDARIES.GEO_OPENSTREETMAP."PLANET_WAYS" w,
        GEO_OPENSTREETMAP_BOUNDARIES.GEO_US_BOUNDARIES."STATES"      s,
        LATERAL FLATTEN(input => w."all_tags")                       tag
WHERE   s."state" = 'CA'
  AND   ST_INTERSECTS( TO_GEOGRAPHY(w."geometry"), TO_GEOGRAPHY(s."state_geom") )
  AND   tag.value:"key"::STRING  =  'highway'
  AND   tag.value:"value"::STRING IN ('motorway','trunk','primary','secondary','residential')
  AND   NOT EXISTS ( SELECT 1
                     FROM   LATERAL FLATTEN(input => w."all_tags") br
                     WHERE  br.value:"key"::STRING = 'bridge')
LIMIT 20;
[Try to execute]
[Error occurred]
{'status': 'error', 'error_msg': '##ERROR##002031 (42601): SQL compilation error:\nUnsupported subquery type cannot be evaluated'}
[Error occurred]
[Corrected SQL]
--Description:  Sample California road "ways" that (1) intersect the California state geometry,
--              (2) possess a highway tag equal to one of the target classes, and
--              (3) have NO tag whose key is 'bridge'.  
--              The logic uses two non-nested FLATTEN() sub-queries (EXISTS / NOT EXISTS) 
--              to bypass Snowflake restriction that caused the earlier compilation error.
SELECT  DISTINCT w."id"
FROM    GEO_OPENSTREETMAP_BOUNDARIES.GEO_OPENSTREETMAP."PLANET_WAYS"  w
JOIN    GEO_OPENSTREETMAP_BOUNDARIES.GEO_US_BOUNDARIES."STATES"       s
       ON s."state" = 'CA'
WHERE   ST_INTERSECTS(
          TO_GEOGRAPHY(w."geometry"),
          TO_GEOGRAPHY(s."state_geom")
        )
  AND   EXISTS (                          -- at least one highway tag of interest
        SELECT 1
        FROM   TABLE(FLATTEN(INPUT => w."all_tags")) tags
        WHERE  tags.VALUE:"key"::STRING   = 'highway'
          AND  tags.VALUE:"value"::STRING IN ( 'motorway',
                                               'trunk',
                                               'primary',
                                               'secondary',
                                               'residential' )
  )
  AND   NOT EXISTS (                      -- absolutely no bridge tag
        SELECT 1
        FROM   TABLE(FLATTEN(INPUT => w."all_tags")) br
        WHERE  br.VALUE:"key"::STRING = 'bridge'
  )
LIMIT 20;
[Corrected SQL]
[Results for corrected sql]
{'status': 'error', 'error_msg': '##ERROR##002031 (42601): SQL compilation error:\nUnsupported subquery type cannot be evaluated'}
[Results for corrected sql]
[Corrected SQL]
--Description:  Return a quick sample (20 ids) of OSM ways that lie in California
--              and satisfy TWO tag-based conditions in a single GROUP BY pass:
--              (1) possess at least one highway tag whose value is motorway, trunk,
--                  primary, secondary, or residential;
--              (2) possess NO tag whose key is bridge.
--              We join California's polygon, flatten tags once, then aggregate with
--              conditional MAX() to implement both tests without any nested-FLATTEN
--              sub-queries (avoids Snowflake's 002031 error).

SELECT  w."id"
FROM    GEO_OPENSTREETMAP_BOUNDARIES.GEO_OPENSTREETMAP."PLANET_WAYS"  w
JOIN    GEO_OPENSTREETMAP_BOUNDARIES.GEO_US_BOUNDARIES."STATES"       s
         ON  s."state" = 'CA'
         AND ST_INTERSECTS(
               TO_GEOGRAPHY(w."geometry"),
               TO_GEOGRAPHY(s."state_geom")
             )
     ,  LATERAL FLATTEN( INPUT => w."all_tags")                       tag
GROUP BY w."id"
HAVING  MAX(
          IFF(tag.value:"key"::STRING = 'highway'
              AND tag.value:"value"::STRING IN ('motorway','trunk',
                                                'primary','secondary',
                                                'residential'),
              1,0)
       ) = 1                -- at least one qualifying highway tag
   AND  MAX(
          IFF(tag.value:"key"::STRING = 'bridge', 1, 0)
       ) = 0                -- zero bridge tags
LIMIT 20;
[Corrected SQL]
[Results for corrected sql]
id
287111786
119334084
758604772
819397379
10368868
10395890
10390036
10392818
10391929
10394870
16226291
10399302
10373342
10371510
16231883
16248611
591503742
841932493
10393626
10390788
[Results for corrected sql]
[Corrected other sqls]
--Description: 7) Count all California ways that (a) intersect the state, (b) carry a
--              highway tag in the target list, and (c) contain NO bridge tag.
SELECT COUNT(*) AS "ca_candidate_roads"
FROM (
      SELECT  w."id"
      FROM    GEO_OPENSTREETMAP_BOUNDARIES.GEO_OPENSTREETMAP."PLANET_WAYS"  w
      JOIN    GEO_OPENSTREETMAP_BOUNDARIES.GEO_US_BOUNDARIES."STATES"       s
             ON  s."state" = 'CA'
             AND ST_INTERSECTS(
                   TO_GEOGRAPHY(w."geometry"),
                   TO_GEOGRAPHY(s."state_geom")
                 )
          ,  LATERAL FLATTEN( INPUT => w."all_tags")                        tag
      GROUP BY w."id"
      HAVING MAX(
               IFF(tag.value:"key"::STRING = 'highway'
                   AND tag.value:"value"::STRING IN ('motorway','trunk',
                                                     'primary','secondary',
                                                     'residential'),
                   1,0)
             ) = 1      -- at least one wanted highway tag
         AND MAX(
               IFF(tag.value:"key"::STRING = 'bridge',1,0)
             ) = 0      -- no bridge tag
     ) AS t;

--Description: 8) Produce a sample of overlapping *pairs* of California candidate
--              roads (built from the filtered set defined in query 7).
WITH candidates AS (
     SELECT  w."id",
             TO_GEOGRAPHY(w."geometry") AS "geom"
     FROM    GEO_OPENSTREETMAP_BOUNDARIES.GEO_OPENSTREETMAP."PLANET_WAYS"  w
     JOIN    GEO_OPENSTREETMAP_BOUNDARIES.GEO_US_BOUNDARIES."STATES"       s
            ON  s."state" = 'CA'
            AND ST_INTERSECTS(
                  TO_GEOGRAPHY(w."geometry"),
                  TO_GEOGRAPHY(s."state_geom")
                )
         ,  LATERAL FLATTEN( INPUT => w."all_tags")                        tag
     GROUP BY w."id", w."geometry"
     HAVING MAX(
              IFF(tag.value:"key"::STRING = 'highway'
                  AND tag.value:"value"::STRING IN ('motorway','trunk',
                                                    'primary','secondary',
                                                    'residential'),
                  1,0)
            ) = 1
        AND MAX(
              IFF(tag.value:"key"::STRING = 'bridge',1,0)
            ) = 0
)
SELECT  c1."id"  AS "road_id_1",
        c2."id"  AS "road_id_2"
FROM    candidates  c1
JOIN    candidates  c2
       ON  c1."id" < c2."id"
       AND ST_INTERSECTS(c1."geom", c2."geom")
LIMIT 20;

--Description: 9) From those overlapping pairs, keep only the pairs that share
--              ZERO common node-ids.
WITH candidates AS (   -- same definition as query 8, but include nodes
     SELECT  w."id",
             TO_GEOGRAPHY(w."geometry") AS "geom",
             w."nodes"
     FROM    GEO_OPENSTREETMAP_BOUNDARIES.GEO_OPENSTREETMAP."PLANET_WAYS"  w
     JOIN    GEO_OPENSTREETMAP_BOUNDARIES.GEO_US_BOUNDARIES."STATES"       s
            ON  s."state" = 'CA'
            AND ST_INTERSECTS(
                  TO_GEOGRAPHY(w."geometry"),
                  TO_GEOGRAPHY(s."state_geom")
                )
         ,  LATERAL FLATTEN( INPUT => w."all_tags")                        tag
     GROUP BY w."id", w."geometry", w."nodes"
     HAVING MAX(
              IFF(tag.value:"key"::STRING = 'highway'
                  AND tag.value:"value"::STRING IN ('motorway','trunk',
                                                    'primary','secondary',
                                                    'residential'),
                  1,0)
            ) = 1
        AND MAX(
              IFF(tag.value:"key"::STRING = 'bridge',1,0)
            ) = 0
),
pairs AS (
     SELECT  c1."id"   AS "road_id_1",
             c1."nodes" AS "nodes1",
             c2."id"   AS "road_id_2",
             c2."nodes" AS "nodes2"
     FROM    candidates c1
     JOIN    candidates c2
            ON c1."id" < c2."id"
           AND ST_INTERSECTS(c1."geom", c2."geom")
)
SELECT  p."road_id_1",
        p."road_id_2"
FROM    pairs p,
        LATERAL FLATTEN( INPUT => p."nodes1") n1,
        LATERAL FLATTEN( INPUT => p."nodes2") n2
GROUP BY p."road_id_1", p."road_id_2"
HAVING  MAX( IFF( n1.value:"id"::NUMBER = n2.value:"id"::NUMBER, 1, 0) ) = 0
LIMIT 20;

--Description: 10) Count all distinct overlapping pairs (from query 9 logic) that
--               have no shared nodes.
WITH candidates AS (
     SELECT  w."id",
             TO_GEOGRAPHY(w."geometry") AS "geom",
             w."nodes"
     FROM    GEO_OPENSTREETMAP_BOUNDARIES.GEO_OPENSTREETMAP."PLANET_WAYS"  w
     JOIN    GEO_OPENSTREETMAP_BOUNDARIES.GEO_US_BOUNDARIES."STATES"       s
            ON  s."state" = 'CA'
            AND ST_INTERSECTS(
                  TO_GEOGRAPHY(w."geometry"),
                  TO_GEOGRAPHY(s."state_geom")
                )
         ,  LATERAL FLATTEN( INPUT => w."all_tags") tag
     GROUP BY w."id", w."geometry", w."nodes"
     HAVING MAX(
              IFF(tag.value:"key"::STRING = 'highway'
                  AND tag.value:"value"::STRING IN ('motorway','trunk',
                                                    'primary','secondary',
                                                    'residential'),
                  1,0)
            ) = 1
        AND MAX(
              IFF(tag.value:"key"::STRING = 'bridge',1,0)
            ) = 0
),
pairs AS (
     SELECT  c1."id"     AS "road_id_1",
             c1."nodes"  AS "nodes1",
             c2."id"     AS "road_id_2",
             c2."nodes"  AS "nodes2"
     FROM    candidates c1
     JOIN    candidates c2
            ON  c1."id" < c2."id"
           AND ST_INTERSECTS(c1."geom", c2."geom")
),
uniq_pairs AS (
     SELECT  p."road_id_1",
             p."road_id_2"
     FROM    pairs p,
             LATERAL FLATTEN(INPUT => p."nodes1") n1,
             LATERAL FLATTEN(INPUT => p."nodes2") n2
     GROUP BY p."road_id_1", p."road_id_2"
     HAVING  MAX( IFF( n1.value:"id"::NUMBER = n2.value:"id"::NUMBER, 1, 0) ) = 0
)
SELECT COUNT(*) AS "total_overlapping_pairs_in_CA"
FROM   uniq_pairs;
[Corrected other sqls]
[Successfully corrected]
id
287111786
119334084
758604772
819397379
10368868
10395890
10390036
10392818
10391929
10394870
16226291
10399302
10373342
10371510
16231883
16248611
591503742
841932493
10393626
10390788
[Successfully corrected]
[Try to execute]
--Description: 7) Count all California ways that (a) intersect the state, (b) carry a
--              highway tag in the target list, and (c) contain NO bridge tag.
SELECT COUNT(*) AS "ca_candidate_roads"
FROM (
      SELECT  w."id"
      FROM    GEO_OPENSTREETMAP_BOUNDARIES.GEO_OPENSTREETMAP."PLANET_WAYS"  w
      JOIN    GEO_OPENSTREETMAP_BOUNDARIES.GEO_US_BOUNDARIES."STATES"       s
             ON  s."state" = 'CA'
             AND ST_INTERSECTS(
                   TO_GEOGRAPHY(w."geometry"),
                   TO_GEOGRAPHY(s."state_geom")
                 )
          ,  LATERAL FLATTEN( INPUT => w."all_tags")                        tag
      GROUP BY w."id"
      HAVING MAX(
               IFF(tag.value:"key"::STRING = 'highway'
                   AND tag.value:"value"::STRING IN ('motorway','trunk',
                                                     'primary','secondary',
                                                     'residential'),
                   1,0)
             ) = 1      -- at least one wanted highway tag
         AND MAX(
               IFF(tag.value:"key"::STRING = 'bridge',1,0)
             ) = 0      -- no bridge tag
     ) AS t
[Try to execute]
[Successfully executed]
ca_candidate_roads
15429
[Successfully executed]
[Try to execute]
--Description: 8) Produce a sample of overlapping *pairs* of California candidate
--              roads (built from the filtered set defined in query 7).
WITH candidates AS (
     SELECT  w."id",
             TO_GEOGRAPHY(w."geometry") AS "geom"
     FROM    GEO_OPENSTREETMAP_BOUNDARIES.GEO_OPENSTREETMAP."PLANET_WAYS"  w
     JOIN    GEO_OPENSTREETMAP_BOUNDARIES.GEO_US_BOUNDARIES."STATES"       s
            ON  s."state" = 'CA'
            AND ST_INTERSECTS(
                  TO_GEOGRAPHY(w."geometry"),
                  TO_GEOGRAPHY(s."state_geom")
                )
         ,  LATERAL FLATTEN( INPUT => w."all_tags")                        tag
     GROUP BY w."id", w."geometry"
     HAVING MAX(
              IFF(tag.value:"key"::STRING = 'highway'
                  AND tag.value:"value"::STRING IN ('motorway','trunk',
                                                    'primary','secondary',
                                                    'residential'),
                  1,0)
            ) = 1
        AND MAX(
              IFF(tag.value:"key"::STRING = 'bridge',1,0)
            ) = 0
)
SELECT  c1."id"  AS "road_id_1",
        c2."id"  AS "road_id_2"
FROM    candidates  c1
JOIN    candidates  c2
       ON  c1."id" < c2."id"
       AND ST_INTERSECTS(c1."geom", c2."geom")
LIMIT 20
[Try to execute]
[Successfully executed]
road_id_1,road_id_2
10392115,10394223
39992298,106472765
39992298,120880124
39992298,39992304
39992298,106472766
39992298,39992302
206509295,206511962
10417068,10426780
10417068,10423563
10648012,196503758
10648012,10676457
16264496,901276711
16264496,166800171
16264496,901276710
10577152,220477764
249251795,249251797
249251795,249251810
249251812,249251813
10577132,364026043
10399110,10399149
[Successfully executed]
[Try to execute]
--Description: 9) From those overlapping pairs, keep only the pairs that share
--              ZERO common node-ids.
WITH candidates AS (   -- same definition as query 8, but include nodes
     SELECT  w."id",
             TO_GEOGRAPHY(w."geometry") AS "geom",
             w."nodes"
     FROM    GEO_OPENSTREETMAP_BOUNDARIES.GEO_OPENSTREETMAP."PLANET_WAYS"  w
     JOIN    GEO_OPENSTREETMAP_BOUNDARIES.GEO_US_BOUNDARIES."STATES"       s
            ON  s."state" = 'CA'
            AND ST_INTERSECTS(
                  TO_GEOGRAPHY(w."geometry"),
                  TO_GEOGRAPHY(s."state_geom")
                )
         ,  LATERAL FLATTEN( INPUT => w."all_tags")                        tag
     GROUP BY w."id", w."geometry", w."nodes"
     HAVING MAX(
              IFF(tag.value:"key"::STRING = 'highway'
                  AND tag.value:"value"::STRING IN ('motorway','trunk',
                                                    'primary','secondary',
                                                    'residential'),
                  1,0)
            ) = 1
        AND MAX(
              IFF(tag.value:"key"::STRING = 'bridge',1,0)
            ) = 0
),
pairs AS (
     SELECT  c1."id"   AS "road_id_1",
             c1."nodes" AS "nodes1",
             c2."id"   AS "road_id_2",
             c2."nodes" AS "nodes2"
     FROM    candidates c1
     JOIN    candidates c2
            ON c1."id" < c2."id"
           AND ST_INTERSECTS(c1."geom", c2."geom")
)
SELECT  p."road_id_1",
        p."road_id_2"
FROM    pairs p,
        LATERAL FLATTEN( INPUT => p."nodes1") n1,
        LATERAL FLATTEN( INPUT => p."nodes2") n2
GROUP BY p."road_id_1", p."road_id_2"
HAVING  MAX( IFF( n1.value:"id"::NUMBER = n2.value:"id"::NUMBER, 1, 0) ) = 0
LIMIT 20
[Try to execute]
[Successfully executed]
road_id_1,road_id_2
186248373,716005664
207150181,716005662
16235945,716005662
16235945,119237655
33194942,108291872
10412617,10424336
16230803,716005664
186248677,188720933
119237655,207150181
184081700,186248373
186248677,253128480
33194941,108291872
16230803,184081700
[Successfully executed]
[Try to execute]
--Description: 10) Count all distinct overlapping pairs (from query 9 logic) that
--               have no shared nodes.
WITH candidates AS (
     SELECT  w."id",
             TO_GEOGRAPHY(w."geometry") AS "geom",
             w."nodes"
     FROM    GEO_OPENSTREETMAP_BOUNDARIES.GEO_OPENSTREETMAP."PLANET_WAYS"  w
     JOIN    GEO_OPENSTREETMAP_BOUNDARIES.GEO_US_BOUNDARIES."STATES"       s
            ON  s."state" = 'CA'
            AND ST_INTERSECTS(
                  TO_GEOGRAPHY(w."geometry"),
                  TO_GEOGRAPHY(s."state_geom")
                )
         ,  LATERAL FLATTEN( INPUT => w."all_tags") tag
     GROUP BY w."id", w."geometry", w."nodes"
     HAVING MAX(
              IFF(tag.value:"key"::STRING = 'highway'
                  AND tag.value:"value"::STRING IN ('motorway','trunk',
                                                    'primary','secondary',
                                                    'residential'),
                  1,0)
            ) = 1
        AND MAX(
              IFF(tag.value:"key"::STRING = 'bridge',1,0)
            ) = 0
),
pairs AS (
     SELECT  c1."id"     AS "road_id_1",
             c1."nodes"  AS "nodes1",
             c2."id"     AS "road_id_2",
             c2."nodes"  AS "nodes2"
     FROM    candidates c1
     JOIN    candidates c2
            ON  c1."id" < c2."id"
           AND ST_INTERSECTS(c1."geom", c2."geom")
),
uniq_pairs AS (
     SELECT  p."road_id_1",
             p."road_id_2"
     FROM    pairs p,
             LATERAL FLATTEN(INPUT => p."nodes1") n1,
             LATERAL FLATTEN(INPUT => p."nodes2") n2
     GROUP BY p."road_id_1", p."road_id_2"
     HAVING  MAX( IFF( n1.value:"id"::NUMBER = n2.value:"id"::NUMBER, 1, 0) ) = 0
)
SELECT COUNT(*) AS "total_overlapping_pairs_in_CA"
FROM   uniq_pairs
[Try to execute]
[Successfully executed]
total_overlapping_pairs_in_CA
13
[Successfully executed]

[Try to run SQL in self-refine]
/*  Count distinct pairs of California roads (motorway | trunk | primary | 
    secondary | residential)  */

WITH ca_state AS (                       -- California state outline
    SELECT TO_GEOGRAPHY("state_geom") AS geom
    FROM   GEO_OPENSTREETMAP_BOUNDARIES.GEO_US_BOUNDARIES."STATES"
    WHERE  "state" = 'CA'
), --------------------------------------------------------------------
candidates AS (                         -- qualifying road "ways" in CA
    SELECT  w."id",
            TO_GEOGRAPHY(w."geometry") AS geom,
            w."nodes"
    FROM    GEO_OPENSTREETMAP_BOUNDARIES.GEO_OPENSTREETMAP."PLANET_WAYS" w
            JOIN ca_state s
              ON  ST_INTERSECTS( TO_GEOGRAPHY(w."geometry"), s.geom )
            ,   LATERAL FLATTEN( INPUT => w."all_tags") tag
    GROUP BY w."id", w."geometry", w."nodes"
    HAVING  MAX( CASE
                   WHEN tag.value:"key"::STRING = 'highway'
                        AND tag.value:"value"::STRING IN
                            ('motorway','trunk','primary',
                             'secondary','residential')
                   THEN 1 ELSE 0 END ) = 1          -- has wanted highway tag
        AND MAX( CASE
                   WHEN tag.value:"key"::STRING = 'bridge' THEN 1 ELSE 0
                 END ) = 0                          -- no bridge tag
), --------------------------------------------------------------------
intersecting_pairs AS (                 -- pairs whose geometries overlap
    SELECT  c1."id"        AS road_id_1,
            c1."nodes"     AS nodes1,
            c2."id"        AS road_id_2,
            c2."nodes"     AS nodes2
    FROM    candidates c1
            JOIN candidates c2
              ON  c1."id" < c2."id"
              AND ST_INTERSECTS(c1.geom, c2.geom)
), --------------------------------------------------------------------
uniq_pairs AS (                         -- keep only pairs with ZERO shared nodes
    SELECT  p.road_id_1,
            p.road_id_2
    FROM    intersecting_pairs p,
            LATERAL FLATTEN( INPUT => p.nodes1 ) n1,
            LATERAL FLATTEN( INPUT => p.nodes2 ) n2
    GROUP BY p.road_id_1, p.road_id_2
    HAVING  MAX( CASE
                   WHEN n1.value:"id"::NUMBER = n2.value:"id"::NUMBER
                   THEN 1 ELSE 0 END ) = 0
) ---------------------------------------------------------------------
SELECT COUNT(*) AS total_overlapping_road_pairs_CA
FROM   uniq_pairs;
[Try to run SQL in self-refine]
[Results]
TOTAL_OVERLAPPING_ROAD_PAIRS_CA
13
[Results]
\end{lstlisting}

\section{Broader Impact}
\label{appendix:broader_impact}

Our work aims to improve the usability and accuracy of natural language interfaces to structured databases through the ReFoRCE framework. By enhancing Text-to-SQL systems with agentic reasoning, schema exploration, and self-refinement, we aim to assist programmers in verifying and refining generated SQL while also democratizing access to data analysis for non-technical users, thereby supporting more inclusive data-driven decision-making in domains such as healthcare, education, public policy, and small business analytics.

However, we also recognize several potential risks. First, incorrect SQL generation in high-stakes domains (e.g., finance, healthcare) may lead to erroneous conclusions or decisions. Second, automated access to databases raises privacy and security concerns, especially if deployed without proper authentication, authorization, and auditing mechanisms. Third, the ability to generate complex queries with minimal oversight could be misused for unauthorized data extraction or inference attacks.

To mitigate these risks, we recommend that any deployment of our system include safeguards such as access controls, human-in-the-loop validation, query logging, and fine-tuned model usage tailored to the domain context. We also encourage further research on robustness, explainability, and error detection for agent-based query generation systems.

\section{Spider2.0 Leaderboard Screenshot}

\begin{figure}
    \centering
    \includegraphics[width=1\linewidth]{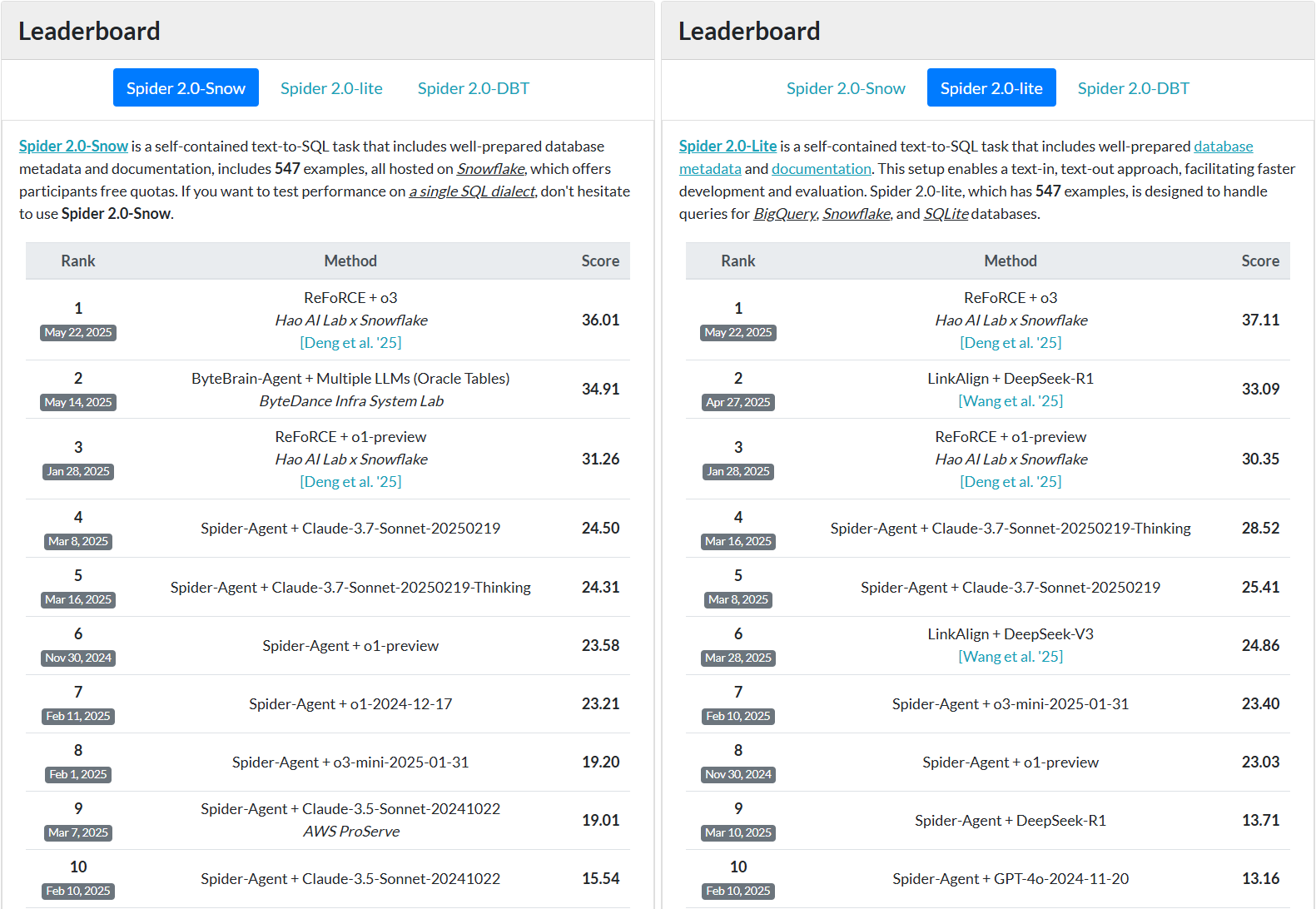}
    \caption{Leaderboard as of May 22, 2025}
\end{figure}